\documentclass{article}
\usepackage{iclr2027_conference,times}
\usepackage[T1]{fontenc}

\usepackage{amsmath,amsfonts,bm}

\newcommand{\obs}{\bm{o}}
\newcommand{\act}{\bm{a}}
\newcommand{\latent}{\bm{z}}
\newcommand{\chunk}{\bm{A}}
\newcommand{\intent}{\bm{m}}
\newcommand{\actionembed}{\bm{u}}
\newcommand{\prevaction}{\bm{b}}
\newcommand{\contextvec}{\bm{c}}
\newcommand{\encoder}{E_{\theta}}
\newcommand{\predictor}{F_{\phi}}
\newcommand{\actionencoder}{A_{\omega}^{\mathrm{VL}}}
\newcommand{\actornet}{G_{\psi}^{\mathrm{AR}}}
\newcommand{\actor}{\pi_{\psi}^{\mathrm{AR}}}
\newcommand{\actormean}{\bm{\mu}_{\psi}}
\newcommand{\actorstd}{\bm{\sigma}_{\psi}}
\newcommand{\residual}{\bm{\epsilon}}

\def\eqref#1{equation~\ref{#1}}

\def\1{\bm{1}}

\DeclareMathAlphabet{\mathsfit}{\encodingdefault}{\sfdefault}{m}{sl}
\SetMathAlphabet{\mathsfit}{bold}{\encodingdefault}{\sfdefault}{bx}{n}

\usepackage{url}
\usepackage{float}
\usepackage{placeins}
\usepackage{multirow}
\usepackage{longtable}
\usepackage{booktabs}
\usepackage{amssymb}
\usepackage{graphicx}
\usepackage{algorithm}
\usepackage{algorithmic}
\usepackage[colorlinks=true,allcolors=blue]{hyperref}
\usepackage{fontawesome5}

\usepackage{xcolor}
\usepackage[normalem]{ulem} %

\title{FlexiWorld: Learning and Planning via Flexible Action Chunks Across Multiple Time Scales}
\author{%
\begin{minipage}{\dimexpr\textwidth-2\tabcolsep\relax}
\raggedright\normalfont
\vspace{4pt}
\mbox{\textbf{Shidu Ren}$^{1*}$}, \mbox{\textbf{Qilin Gu}$^{1*}$},
\mbox{\textbf{Zhenghao Ni}$^{1*}$}, \mbox{\textbf{Junhan Sun}$^{2}$},
\mbox{\textbf{Jiaqi Wang}$^{3}$}, \mbox{\textbf{Damien Scieur}$^{4,5}$},
\mbox{\textbf{Yunze Liu}$^{6\dagger}$}\\[3pt]
\mbox{$^{1}$University of Toronto}, \mbox{$^{2}$Zhejiang University},
\mbox{$^{3}$Tencent Jarvis Lab}, \mbox{$^{4}$Mila \& Universit\'{e} de Montr\'{e}al},
\mbox{$^{5}$Samsung SAIL}, \mbox{$^{6}$Tsinghua University}\\[2pt]
$^{*}$Equal contribution\quad $^{\dagger}$Corresponding author\\[2pt]
{\hypersetup{urlcolor=black}\href{https://shidu-ren.github.io/FlexiWorld-Project-Page/}{\faGlobe\ Project Page}\quad
\href{https://github.com/Shidu-Ren/FlexiWorld}{\faGithub\ Code}\quad
\href{https://huggingface.co/ryanren0330/FlexiWorld}{\raisebox{-1pt}{\includegraphics[height=10pt]{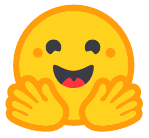}}\ Models}}
\end{minipage}}
\iclrfinalcopy
\hypersetup{pdftitle={FlexiWorld: Learning and Planning via Flexible Action Chunks Across Multiple Time Scales},
pdfauthor={Shidu Ren, Qilin Gu, Zhenghao Ni, Junhan Sun, Jiaqi Wang, Damien Scieur, Yunze Liu}}

\begin{document}
\raggedbottom

\vspace*{-28pt}
\maketitle
\lhead{Preprint}
\vspace{-24pt}

\begin{abstract}
Latent world models predict future states for goal-directed planning using action chunks spanning multiple primitive steps. Existing methods typically use
fixed-length chunks and either omit goal-conditioned action generation or limit
their supervision to short goal spans. We introduce FlexiWorld, a JEPA-based
world model that combines mixed-span goal supervision with variable-length
action chunks to improve long-horizon control. During training, we sample varying goal spans and randomly partition the actions into variable-length chunks.
We jointly train the world model with a causal action encoder that
embeds variable-length chunks and an autoregressive actor that generates primitive actions sequentially. Student Forcing reduces exposure bias by training on generated action prefixes. For planning, Actor-Residual Cross-Entropy Method
(ARCEM) combines action-residual search with within-chunk autoregressive feedback and chunk-boundary latent prediction. Across four benchmarks and goal distances,
FlexiWorld with ARCEM achieves 89.29\% mean success, compared with 83.98\%
for the strongest baseline. PushT ablations show improved
direct control from mixed-span supervision, variable-length chunks, and
Student Forcing. Without retraining, FlexiWorld supports different planning
chunk lengths: longer chunks accelerate ARCEM by approximately $1.3\times$ on average
while maintaining comparable average success.
\end{abstract}
\vspace{-8pt}
\begingroup
\setlength{\intextsep}{6pt}
\setlength{\abovecaptionskip}{4pt}
\begin{figure}[H]
\centering
\includegraphics[width=\linewidth]{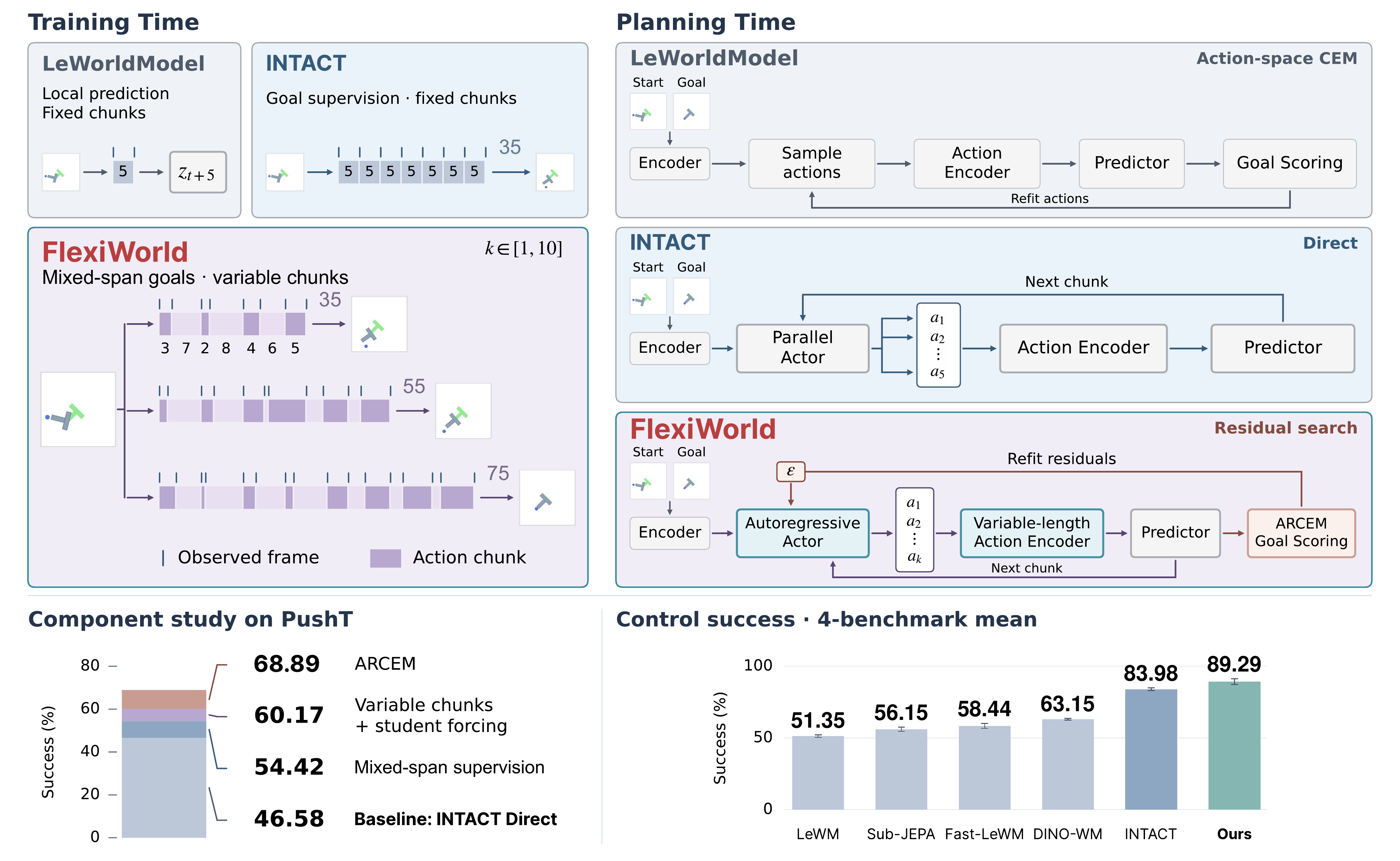}
\caption{\textbf{FlexiWorld: flexible action chunks across multiple time scales.}
Training and planning interfaces (top), four-benchmark mean success
(bottom right), and selected PushT configurations (bottom left).
Tables~\ref{tab:main-results} and~\ref{tab:ablations-extended} detail the comparisons.}
\label{fig:planning-comparison}
\end{figure}
\endgroup
\section{Introduction}

Latent world models allow an agent to plan from visual observations by
predicting future states before executing actions. World models
based on Joint Embedding Predictive Architectures (JEPAs) make these
predictions in a learned representation space, without reconstructing
future images \citep{assran2023ijepa,assran2025vjepa2,maes2026leworldmodel}.
By grouping primitive actions into action chunks, a model can predict
the state reached after several environment steps in a single transition.
A planner selects actions by comparing predicted and goal latent states.
Reaching distant goals
therefore requires both goal-conditioned action generation and latent
prediction over multiple time scales.

Prior work on JEPA-based world models has explored learned dynamics,
goal-conditioned action generation, and longer-horizon prediction.
LeWorldModel (LeWM) \citep{maes2026leworldmodel}
predicts local latent transitions and uses the Cross-Entropy Method (CEM)
to search for actions at test time. INTACT \citep{sun2026intact} trains a shared actor with local
inverse-dynamics and future-goal objectives, enabling action generation
without search. VLWM \citep{du2026vlwm} and
Fast-LeWM \citep{gao2026fastleworldmodel} extend prediction across multiple
action chunks through variable-horizon and parallel prefix prediction,
respectively.
VLWM feeds action tokens to a shared predictor and gradually expands its
training horizon, allowing a single prediction to span multiple fixed action chunks.
Fast-LeWM uses a causal action-prefix encoder to predict multiple future
states directly from the same observed latent state, avoiding a sequential
chain of latent predictions.

Despite these advances, limitations remain in goal-conditioned action
generation and action chunking. LeWM relies on search without learning how to
generate actions, while INTACT limits goal supervision to short spans,
leaving its actor without direct supervision for more distant goals.
Both use fixed-length
action chunks, so each predicted transition spans the same number of
primitive steps. This fixes the temporal granularity of their transitions,
preventing planners from adjusting chunk length to trade predictor calls
against temporal resolution. VLWM and Fast-LeWM extend prediction across
multiple action chunks while retaining fixed-size chunks. Their flexibility therefore concerns how
many action chunks a prediction spans, rather than the primitive-action
boundaries of those chunks. Both use CEM without jointly
training a goal-conditioned actor, so long-horizon prediction does not
directly supervise action generation. Neither combines mixed-span goal
supervision with flexible chunking, limiting distant-goal action learning
and planning granularity.

We introduce FlexiWorld, a JEPA-based world model that combines
mixed-span goal supervision with variable-length action chunks to
improve long-horizon control (Figure~\ref{fig:planning-comparison}).
During training, we vary goal spans and randomly partition trajectories
into variable-length action chunks, jointly supervising latent prediction
and goal-conditioned action generation across time scales.
A causal action encoder represents chunks of different lengths, and an
autoregressive actor generates the requested number of primitive actions
conditioned on the current latent state and a local or distant goal.
Teacher Forcing trains this actor on expert prefixes, whereas
planning uses its own generated prefixes, creating exposure bias.
We address this mismatch with Student Forcing (SF)
\citep{bengio2015scheduledsampling}, which trains on both expert and
self-generated prefixes and improves success in our component study.
We jointly train the action encoder and actor with the world model, sharing each
module's parameters across chunk lengths.

FlexiWorld's autoregressive policy supports search-free Direct planning,
generating each chunk from the predicted latent state.
POPLIN~\citep{wang2019poplin} refines policy outputs with action
residuals, predicting a new state before recomputing each subsequent action.
We introduce Actor-Residual Cross-Entropy Method (ARCEM), which generalizes
this search to autoregressive chunks: perturbed actions condition subsequent
outputs within a chunk, while latent states are predicted only at chunk
boundaries.

We evaluate FlexiWorld on PushT, OGBench-Cube (Cube), Reacher, and
TwoRoom across multiple goal distances.
FlexiWorld achieves 89.29\% mean success with ARCEM's test-time search
versus 83.98\% for INTACT's strongest configuration, and 86.79\% with
Direct action generation versus 81.62\% for INTACT Direct.
A PushT component study finds little change from the architecture alone,
but gains from mixed-span goal supervision. Variable-length chunks and
Student Forcing also improve success when evaluated separately.
Distant-goal action-prediction tests also favor FlexiWorld, while CEM
without either actor gives similar success for both models, suggesting
the clearest benefit is in goal-conditioned action generation.

The same trained model supports longer action chunks to reduce predictor
calls at planning time. Using ten-action rather than five-action chunks
makes ARCEM approximately $1.3\times$ faster at 50- and 100-step goals,
with comparable four-benchmark mean success and task-dependent gains and
losses. Longer chunks also reduce expert-action latent rollout error on
PushT and Cube at both distances, although lower error does not
consistently improve control. Section~\ref{sec:analysis} analyzes these
deployment trade-offs.

Our contributions are:
\textbf{(i)} We introduce FlexiWorld, jointly learning latent prediction and
autoregressive action generation across variable-length chunks with
mixed-span goal supervision.
\textbf{(ii)} We develop ARCEM, generalizing POPLIN's action-residual
search to autoregressive chunks: perturbed actions condition subsequent
generation, with latent prediction only at chunk boundaries.
\textbf{(iii)} We benchmark FlexiWorld against JEPA-based baselines under a
unified evaluation protocol across four tasks and multiple goal distances,
demonstrating improved goal-reaching success.

\section{Related Work}

\textbf{Latent world models for goal-conditioned control.}
Latent world models learn dynamics for control \citep{ha2018worldmodels}.
PlaNet and Dreamer use image reconstruction
\citep{hafner2019planet,hafner2019dreamer,hafner2023dreamerv3}, whereas
TD-MPC combines latent prediction with value learning
\citep{hansen2022tdmpc,hansen2023tdmpc2}. JEPA-based models predict
representations without image reconstruction
\citep{bardes2024vjepa,assran2025vjepa2,maes2026leworldmodel}.
DINO-WM \citep{zhou2024dinowm} uses pre-trained visual features, whereas
LeWM \citep{maes2026leworldmodel} and Sub-JEPA \citep{zhao2026subjepa}
jointly learn representations and dynamics through Gaussian regularization.
GC-IDM learns control from world-model features \citep{nguyen2026latentgeometry}.
Qantara \citep{rakhimov2026qantara} connects planning, action sampling, and
inverse dynamics through bridge-flow training. INTACT \citep{sun2026intact}
jointly learns prediction and a shared intent-to-action model for search-free
control. Its action encoder and actor use fixed-length chunks, keeping local
transition supervision at a fixed duration even when goal spans increase.
FlexiWorld varies both transition durations and goal spans.

\textbf{Temporal abstraction and action chunking.}
Action chunking sets the temporal granularity of prediction and control.
HWM \citep{zhang2026hwm} learns variable-duration macro-actions for
hierarchical planning, VLWM \citep{du2026vlwm} varies prediction horizons,
and Fast-LeWM \citep{gao2026fastleworldmodel} predicts action-prefix
outcomes in parallel. In their reported implementations, VLWM and Fast-LeWM
retain fixed base action blocks. On the policy side, ACT \citep{zhao2023act}
learns action chunks, while ARP \citep{zhang2025arp} combines autoregression
with chunked prediction. BID \citep{liu2025bidirectional} uses guided
resampling to balance temporal coherence with reactivity. Adaptive chunking
selects execution lengths using action entropy \citep{liang2026adaptivechunking}
or values learned through offline-to-online reinforcement learning
\citep{shin2026adaptivechunking}. These approaches either learn multi-scale
latent dynamics for search without jointly training a goal-conditioned
actor, or organize action generation and execution without jointly learning
variable-duration latent dynamics. FlexiWorld instead varies chunk boundaries
at primitive-action resolution and jointly trains its action encoder,
autoregressive actor, and latent predictor over the resulting chunks.
Combined with mixed-span goal supervision, this supports goal-conditioned
action generation and latent prediction at adjustable temporal granularities.

\textbf{Policy-guided model predictive control.}
PETS \citep{chua2018pets} and iCEM \citep{pinneri2020icem} optimize action
sequences using learned dynamics; policies can guide such search by proposing
or adapting candidate plans.
POPLIN \citep{wang2019poplin} explores action-residual and policy-parameter
search. Its fixed-rollout variant perturbs a policy rollout, whereas its
replanning variant recomputes policy outputs along each perturbed state
trajectory. PRISM \citep{wang2026prism} fuses a
state-and-goal-conditioned Gaussian action prior with the planner's initial
distribution through a product of Gaussians. INTACT
\citep{sun2026intact} preserves a Direct reference during Guarded-A's
local action-space search, without regenerating the actor's outputs.
ARCEM generalizes POPLIN's action-residual replanning to autoregressive action
chunks: each perturbed action conditions subsequent outputs within a chunk,
while latent predictions carry its effects across chunks.
Earlier residuals thus change the conditional means around which later
residuals are applied.
Unlike residual RL, which learns controller corrections
\citep{johannink2018residual}, ARCEM optimizes action residuals at test time
without training a residual policy.

\section{Methodology}
\subsection{Method Overview}
\label{sec:method-overview}

\begin{figure}[t]
\centering
\includegraphics[width=\linewidth]{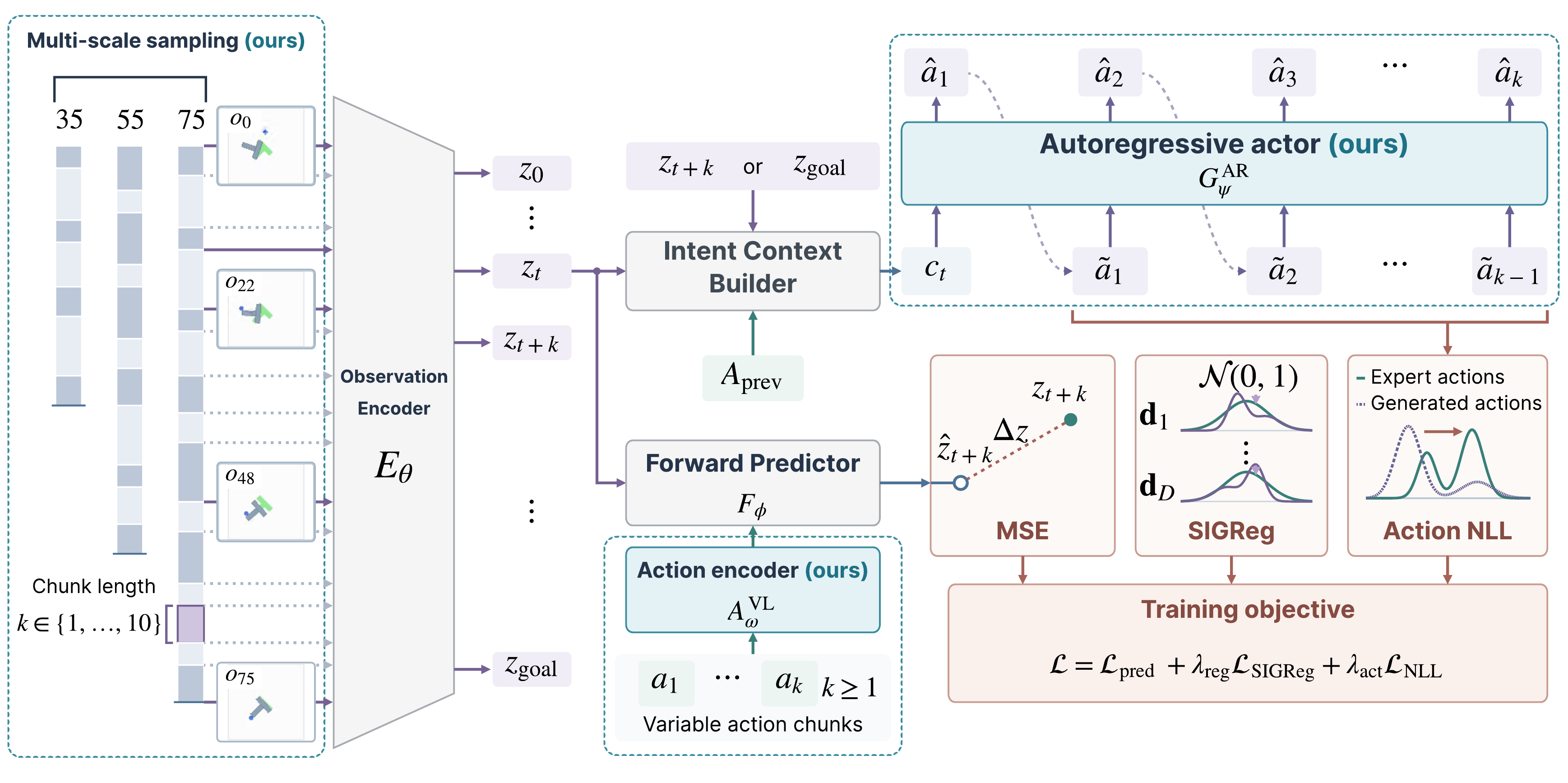}
\caption{\textbf{Joint prediction and action learning in FlexiWorld.}
Training samples different goal spans and partitions each window into
variable-length action chunks paired with their boundary observations.
Chunk embeddings condition the predictor on the actions leading to the
next boundary latent state. The actor is supervised by expert actions,
conditioned on local or final-goal intents and expert or generated action
prefixes. Training combines prediction MSE, SIGReg, and action negative
log-likelihood (NLL).
All modules, including the observation encoder and predictor, are trained
jointly.}
\label{fig:odyssey-training}
\end{figure}

FlexiWorld jointly learns latent prediction and goal-conditioned action
generation from offline trajectories. Figure~\ref{fig:odyssey-training}
shows how training windows with different goal spans are partitioned into
variable-length action chunks. The action encoder embeds each chunk, and
the predictor combines this embedding with the latent-state history to
predict the next boundary latent state. The actor learns to generate the
chunk's actions conditioned on the current latent state and either the
next boundary observation or the final goal, expressed through latent-state
differences.
At deployment, the actor and predictor alternate to construct a Direct plan
with chosen chunk lengths, which ARCEM can refine through action residuals.

\subsection{Multi-Time-Scale Training}
\label{sec:multi-time-scale}

FlexiWorld varies both the training goal span and the action chunk length
to learn from goals at different temporal distances and transitions of
different durations. Given an offline trajectory of pixel observations
$\obs_t$ and primitive actions $\act_t\in\mathbb R^{d_a}$, we sample a
window starting at environment step $s$ with goal span $S\in\mathcal S$,
where $\mathcal S$ is a finite set of spans. In our experiments,
$\mathcal S=\{35,55,75\}$ primitive steps. The window starts with
observation $\obs_s$ and uses the recorded observation $\obs_{s+S}$,
$S$ primitive steps later, as its goal. Within this window,
we partition the actions into $N=N(S)$ chunks, using $N=7,11,15$
for the respective spans, so the mean chunk length remains five.
Each chunk length $k_i$ sets the duration of
one supervised transition. To vary these durations while keeping the
window endpoints fixed, we sample the length sequence uniformly from
\begin{equation}
 \mathcal K_{S,N}=\left\{(k_0,\ldots,k_{N-1})\in\mathbb Z^N\colon
 k_{\min}\leq k_i\leq k_{\max},\quad
 \sum_i k_i=S\right\}\setminus\{(S/N,\ldots,S/N)\}.
 \label{eq:partitions}
\end{equation}
This excludes partitions in which every chunk has the same length.
The sampled lengths define boundaries $t_i=s+\sum_{r<i}k_r$ for
$i=0,\ldots,N$. Each action chunk
$\chunk_i=(\act_{t_i},\ldots,\act_{t_i+k_i-1})$, together with its starting
and ending observations $\obs_{t_i}$ and $\obs_{t_{i+1}}$, forms a supervised
transition. Different partitions therefore provide different intermediate
transitions while sharing the same final goal $\obs_{s+S}$.
Appendix~\ref{app:sampling} gives the sampling implementation.

\subsection{Flexible Action Chunks}
\label{sec:flexible-action-chunks}

To learn from the variable-duration transitions sampled above, FlexiWorld
pairs a variable-length action encoder with an autoregressive actor,
jointly trained with latent prediction and Student Forcing, described below
under joint training.

\textbf{Action encoding.}
The action encoder maps each sampled chunk to a fixed-dimensional embedding
for the predictor. Following LeWM~\citep{maes2026leworldmodel}, the
observation encoder gives $\latent_i=\encoder(\obs_{t_i})$ at each chunk
boundary. Our variable-length (VL) causal Transformer action encoder
$\actionencoder$ processes $\chunk_i$
with positional encodings~\citep{vaswani2017attention}, reads its last valid
hidden state, and adds a learned chunk-length embedding. The resulting
vector $\actionembed_i=\actionencoder(\chunk_i,k_i)$ conditions the predictor
$\hat{\latent}_{i+1}=\predictor(\mathcal H_i,\actionembed_i)$, where
$\mathcal H_i$ contains the history of encoded observations and expert
chunks during training. The same action encoder supplies the actor's
previous-chunk context $\prevaction_i=\actionencoder(\chunk_{i-1},k_{i-1})$;
$\prevaction_0$ encodes the action history before the window.

\textbf{Action generation.}
Our autoregressive (AR) actor $\actornet$ generates primitive actions,
allowing one shared model to produce chunks of different lengths. We retain INTACT's
conditioning scheme~\citep{sun2026intact}, with local intent
$\intent_i^{\mathrm{local}}=\latent_{i+1}-\latent_i$ and
final-goal intent $\intent_i^{\mathrm{goal}}=\operatorname{sg}(\latent_N)-\latent_i$,
where $\operatorname{sg}$ stops gradients at the final-goal occurrence.
For either intent $q\in\{\mathrm{local},\mathrm{goal}\}$, the Intent Context
Builder in Figure~\ref{fig:odyssey-training} forms the actor context
$\contextvec_i^q=[\latent_i;\intent_i^q;\latent_i\odot\intent_i^q;\prevaction_i]$,
where $\odot$ denotes elementwise multiplication.
Unlike INTACT's fixed-width action output, $\actornet$ maps the context
and within-chunk prefix to a Gaussian mean $\actormean$ and standard
deviation $\actorstd$. Its conditional action distribution $\actor$
factorizes over primitive actions:
\begin{equation}
 \actor(\chunk_i\mid\contextvec_i^q)
 =\prod_{j=0}^{k_i-1}
 \actor(\act_{t_i+j}\mid\contextvec_i^q,\chunk_{i<j}),
 \label{eq:primitive-actor}
\end{equation}
where $\chunk_{i<j}$ contains the first $j$ actions and each conditional is
a diagonal Gaussian parameterized by $\actornet$.
The length $k_i$ sets the decoding steps, requiring no length token or
length-specific head. Encoding executable actions ties predictions to
concrete candidate sequences without a separate latent-action decoder.

\textbf{Joint training.}
We jointly supervise latent prediction and action generation on the sampled
chunks. During planning, the actor conditions on its own earlier actions,
which can differ from the expert prefixes used in standard teacher-forced
training. Student Forcing addresses this mismatch by mixing expert
and generated prefixes, following scheduled sampling~\citep{bengio2015scheduledsampling}.
For each chunk and intent, we greedily decode $\hat{\chunk}_i^q$ from
$\contextvec_i^q$ and make one sequence-level choice for all within-chunk prefixes:
\begin{samepage}
\begin{equation}
 \widetilde{\chunk}_i^q=
 \begin{cases}
  \operatorname{sg}(\hat{\chunk}_i^q), & \text{with probability }p_{\mathrm{SF}},\\
  \chunk_i, & \text{otherwise}.
 \end{cases}
 \label{eq:student-prefix}
\end{equation}
At position $j$, only the first $j$ actions of this sequence are supplied,
while the target remains expert action $\act_{t_i+j}$:
\begin{equation}
 \ell_i^q=-\frac{1}{k_i d_a}\sum_{j=0}^{k_i-1}
 \log \actor\!\left(\act_{t_i+j}\mid\contextvec_i^q,
 \widetilde{\chunk}_{i<j}^{q}\right).
 \label{eq:chunk-action-loss}
\end{equation}
\end{samepage}
The context $\contextvec_i^q$ stays fixed across prefix choices, with no
gradient through generated prefixes. Normalization gives chunks equal
weight regardless of length. Averaging over chunks and samples with
fixed relative weights for the two intents gives $\mathcal L_{\mathrm{NLL}}$.
We train all modules from random initialization with
\begin{equation}
 \mathcal L=\mathcal L_{\mathrm{pred}}
 +\lambda_{\mathrm{act}}\mathcal L_{\mathrm{NLL}}
 +\lambda_{\mathrm{reg}}\mathcal L_{\mathrm{SIGReg}}.
 \label{eq:odyssey-loss}
\end{equation}
Here $\mathcal L_{\mathrm{pred}}$ is the mean-squared latent prediction loss,
which updates both prediction and target branches. The
Sketched-Isotropic-Gaussian Regularizer (SIGReg)
\citep{balestriero2025lejepa,maes2026leworldmodel} regularizes boundary latents
toward an isotropic Gaussian. Appendix~\ref{app:training} specifies the
loss weights, averaging, and two-pass Student Forcing implementation.

\begin{figure}[t]
\centering
\includegraphics[width=\linewidth]{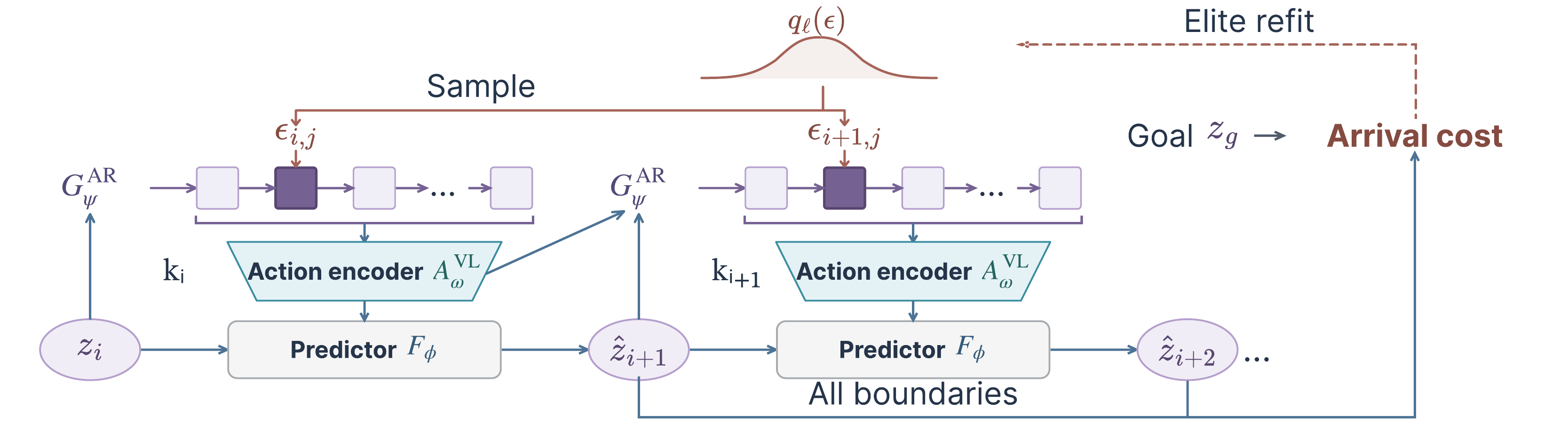}
\caption{\textbf{ARCEM refinement.} Residuals modify actions and subsequent
actor conditions. Chunk embeddings and predicted states propagate these
changes; arrival costs select elites to refit the residual distribution.}
\label{fig:odyssey-arcem}
\end{figure}

\subsection{Direct Planning and Actor-Residual CEM}
\label{sec:ar-cem}

\textbf{Direct planning.}
FlexiWorld generates plans at a chosen chunk length without search or
retraining. Following INTACT~\citep{sun2026intact}, we initialize
$\hat{\latent}_0=\encoder(\obs_s)$ and $\latent_g=\encoder(\obs_g)$,
then alternate the actor and predictor. At boundary $i$, $\actornet$
decodes $k_i$ conditional means using goal intent
$\latent_g-\hat{\latent}_i$ and preceding-chunk context. The chunk embedding
$\actionencoder(\chunk_i,k_i)$ conditions both $\predictor$'s next-state
prediction and the next actor call. For a $D$-action plan,
$\sum_{i=0}^{H-1}k_i=D$ over $H$ predicted transitions. Choosing five-action
or ten-action chunks, with a shorter final chunk when needed, changes
the rollout resolution: longer chunks require fewer predictor calls but
still generate all $D$ actions autoregressively. Reobservation timing is
independent of chunk length (Section~\ref{sec:experimental-setup}).

\textbf{Actor-residual CEM.}
The Cross-Entropy Method (CEM)
\citep{rubinstein1999cem,chua2018pets,pinneri2020icem} iteratively samples plans
and refits its distribution to low-cost elites.
ARCEM generalizes POPLIN's action-residual recursion
\citep{wang2019poplin} to autoregressive chunks (Figure~\ref{fig:odyssey-arcem}).
POPLIN updates states after each perturbed action; ARCEM uses
within-chunk prefix feedback and updates latents at boundaries. At $k=1$,
candidate generation recovers POPLIN with FlexiWorld's actor and
dynamics, while ARCEM retains its own objective and CEM settings.
Chunks require $\lceil D/k\rceil$ predictor calls per $D$ actions, without
within-chunk state updates.
At position $j$ of chunk $i$, the candidate's predicted latent state,
goal intent, preceding-chunk embedding, and generated prefix form
$\contextvec_{i,j}$. We perturb the actor mean with an action residual:
\begin{equation}
 \act_{i,j}=\actormean(\contextvec_{i,j})+
 T\,\residual_{i,j}.
\label{eq:residual-decoding}
\end{equation}
With all residuals set to zero, this recursion recovers the Direct plan
for the same initial context and chunk schedule.
Changing chunk length adjusts the frequency of latent prediction while
retaining a residual for every primitive action. Temperature $T$ scales residuals in
normalized action coordinates, without multiplying by the actor's
predicted standard deviation. We allow predicted arrival before the final chunk by
using the closest boundary to the goal:
\begin{equation}
 \mathcal C(\residual)=\min_{1\leq i\leq H}
 \|\hat{\latent}_i(\residual)-\latent_g\|_2^2,
 \label{eq:arrival-cost}
\end{equation}
where $\residual$ collects all primitive residuals in a candidate plan.
ARCEM starts from a standard Gaussian over residual sequences and updates
its mean and diagonal covariance from the elites. Each iteration samples new candidates
and includes the exact Direct plan, retaining the best candidate across
iterations. The lowest-cost plan is returned using the same trained modules
as Direct, without additional training.

\section{Experiments}
We evaluate whether FlexiWorld improves visual goal-reaching across four
benchmarks, which training components contribute to its performance, and
whether longer action chunks accelerate planning while maintaining comparable
success. We vary planning chunk length using the same trained checkpoints
and use action-prediction, frozen-feature, and paired execution diagnostics
to examine the gains and remaining limitations.

\subsection{Experimental Setup}
\label{sec:experimental-setup}

\textbf{Datasets.}
We use offline expert trajectories from the LeWM benchmark
\citep{maes2026leworldmodel}. Its four visual goal-reaching tasks cover
planar T-block pushing in PushT, 3D manipulation in OGBench-Cube
(Cube; \citealp{park2025ogbench}), arm
configuration matching in Reacher, and navigation through connected rooms in TwoRoom.
Each episode starts from a recorded state. Its goal observation lies
$D\in\{25,50,75,100\}$ primitive steps later in that trajectory; $D$ is the goal distance.

\textbf{Baselines.}
We compare FlexiWorld with LeWM, Fast-LeWM, Sub-JEPA, DINO-WM, and INTACT
\citep{maes2026leworldmodel,gao2026fastleworldmodel,zhao2026subjepa,zhou2024dinowm,sun2026intact}.
LeWM, Fast-LeWM, Sub-JEPA, and DINO-WM provide latent world-model baselines
for action-space planning, with DINO-WM using pre-trained visual features.
INTACT additionally learns goal-conditioned action generation with fixed-length
chunks, providing a direct comparison for search-free control.
Table~\ref{tab:main-results} reports each method's strongest evaluated
configuration by overall mean success, and Table~\ref{tab:planner-comparison}
compares planning variants of INTACT and FlexiWorld.
Baseline configurations and training budgets are detailed in
Appendix~\ref{app:baseline-training}.

\textbf{Metrics.}
Following LeWM, we characterize goal-reaching control by the goal distance
and the execution budget. We report success under each environment's
criterion with a budget of $2D$ primitive steps. The controller executes
a $D$-step plan, then replans once from a new observation if needed.
The evaluation protocol specifies 100 episodes per distance and evaluation
seed in $\{0,1,42\}$. FlexiWorld uses training seeds $\{0,42,3072\}$.
Appendix~\ref{app:statistics} details evaluation coverage and the aggregation
of mean success and sample standard deviations.

\textbf{Implementation details.}
FlexiWorld trains for two mixed-span epochs using 7, 11, or 15 chunks over
goal spans of 35, 55, or 75 primitive steps, respectively, with
$k\in\{1,\ldots,10\}$ and $p_{\mathrm{SF}}=0.5$.
The main comparison uses $k=5$ and $H=D/5$, matching LeWM and INTACT;
Section~\ref{sec:analysis} also evaluates $k=10$ without retraining.
Search budgets (candidates per iteration $\times$ iterations) are
$128\times3$ for ARCEM and Guarded-A.
ARCEM uses 16 elites and $T=0.2$. Appendices~\ref{app:training}
and~\ref{app:search} detail architectures and budgets.

\suppressfloats[t]
\begin{table}[t]
\begingroup
\renewenvironment{table}[1][]{}{}
\begin{table}[t]
\centering
\setlength{\tabcolsep}{3pt}
\caption{\textbf{Four-distance success rate (\%).}
One-replan success averages four distances; Average also averages tasks.
Each method uses its best evaluated configuration: Guarded-A for INTACT,
ARCEM for FlexiWorld. Entries are mean $\pm$ sample standard deviation (SD)
across training seeds for FlexiWorld and evaluation seeds for baselines
(Appendix~\ref{app:statistics}).}
\label{tab:main-results}
\begingroup\small
\begin{tabular*}{\linewidth}{@{\extracolsep{\fill}}lrrrrr@{}}
\toprule
Method & PushT & Cube & Reacher & TwoRoom & Average \\
\midrule
LeWM & $33.83\!\pm\!0.52$ & $51.67\!\pm\!2.31$ & $72.08\!\pm\!1.66$ & $47.83\!\pm\!0.72$ & $51.35\!\pm\!0.88$ \\
Fast-LeWM & $42.42\!\pm\!2.04$ & $55.33\!\pm\!3.82$ & $73.17\!\pm\!1.53$ & $62.83\!\pm\!1.46$ & $58.44\!\pm\!1.67$ \\
Sub-JEPA & $40.58\!\pm\!1.26$ & $54.33\!\pm\!3.76$ & $74.67\!\pm\!1.77$ & $55.00\!\pm\!1.75$ & $56.15\!\pm\!1.51$ \\
DINO-WM & $39.08\!\pm\!0.38$ & $55.83\!\pm\!3.84$ & $66.83\!\pm\!4.38$ & $90.83\!\pm\!0.29$ & $63.15\!\pm\!0.59$ \\
INTACT & $55.17\!\pm\!1.63$ & $84.58\!\pm\!2.18$ & $98.42\!\pm\!0.52$ & $\mathbf{97.75}\!\pm\!\mathbf{0.75}$ & $83.98\!\pm\!0.96$ \\
\midrule
\textbf{FlexiWorld} & $\mathbf{68.89}\!\pm\!\mathbf{4.80}$ & $\mathbf{91.94}\!\pm\!\mathbf{1.50}$ & $\mathbf{99.72}\!\pm\!\mathbf{0.17}$ & $96.61\!\pm\!2.45$ & $\mathbf{89.29}\!\pm\!\mathbf{1.96}$ \\
\bottomrule
\end{tabular*}
\par\endgroup
\end{table}

\par\vspace{0pt}
\begin{table}[t]
\centering\setlength{\tabcolsep}{3pt}
\caption{\textbf{Planning performance of INTACT and FlexiWorld.}
Both models use Direct and Guarded-A; FlexiWorld also uses ARCEM.
Success rates (\%) follow the aggregation and SD
conventions of Table~\ref{tab:main-results}. Appendix~\ref{app:search}
reports search budgets and planning times.}
\label{tab:planner-comparison}
\begingroup\small
\begin{tabular*}{\linewidth}{@{\extracolsep{\fill}}llrrrrr@{}}
\toprule
Model & Planner & PushT & Cube & Reacher & TwoRoom & Average \\
\midrule
\multirow{2}{*}{INTACT}
 & Direct & $46.58\!\pm\!1.81$ & $85.50\!\pm\!2.78$ & $98.58\!\pm\!0.95$ & $95.83\!\pm\!1.44$ & $81.62\!\pm\!0.78$ \\
 & Guarded-A & $55.17\!\pm\!1.63$ & $84.58\!\pm\!2.18$ & $98.42\!\pm\!0.52$ & $\mathbf{97.75}\!\pm\!\mathbf{0.75}$ & $83.98\!\pm\!0.96$ \\
\midrule
\multirow{3}{*}{\textbf{FlexiWorld}}
 & Direct & $60.39\!\pm\!3.92$ & $91.36\!\pm\!1.56$ & $99.06\!\pm\!0.05$ & $96.36\!\pm\!1.92$ & $86.79\!\pm\!1.65$ \\
 & Guarded-A & $65.64\!\pm\!5.22$ & $90.69\!\pm\!2.18$ & $97.83\!\pm\!0.51$ & $97.53\!\pm\!1.89$ & $87.92\!\pm\!2.26$ \\
 & \textbf{ARCEM} & $\mathbf{68.89}\!\pm\!\mathbf{4.80}$ & $\mathbf{91.94}\!\pm\!\mathbf{1.50}$ & $\mathbf{99.72}\!\pm\!\mathbf{0.17}$ & $96.61\!\pm\!2.45$ & $\mathbf{89.29}\!\pm\!\mathbf{1.96}$ \\
\bottomrule
\end{tabular*}
\par\endgroup
\end{table}

\par\vspace{4pt}
\makeatletter\def\@captype{figure}\makeatother
\centering
\includegraphics[width=\linewidth]{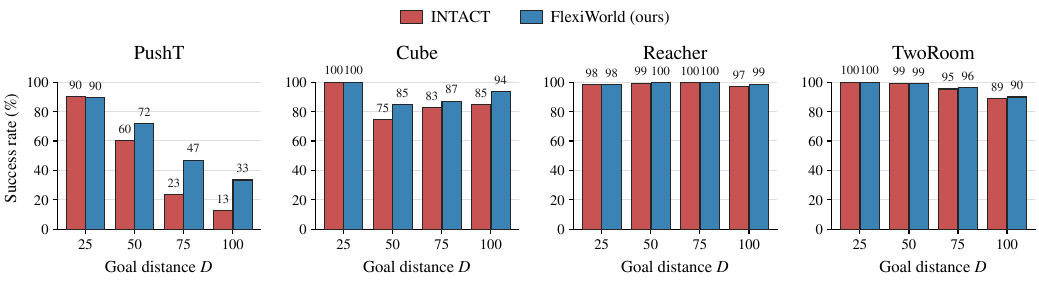}
\caption{\textbf{Direct planning at increasing goal distances.}
Direct success with one replan across four tasks. Goal distance is in
primitive steps; bar labels round success to the nearest percent.}
\label{fig:distance-results}
\endgroup
\end{table}

\subsection{Main Results}
\label{sec:main-results}

\begin{table}[t]
\centering\setlength{\tabcolsep}{3pt}
\caption{\textbf{PushT component study.} We compare INTACT baselines with
our encoder and actor variants. Direct success averages four distances with
up to one replan; results are mean $\pm$ SD across three evaluation seeds
(training seed 0). $\checkmark$/-- mark enabled/disabled; SF denotes Student
Forcing ($p_{\mathrm{SF}}=0.5$). Mixed spans use 35/55/75-step goals for two
epochs; single spans use 35-step goals (75 where indicated) for six.
Appendix~\ref{app:sampling} details evaluation provenance.}
\label{tab:ablations}
\begingroup\small
\begin{tabular*}{\linewidth}{@{\extracolsep{\fill}}lcccr@{}}
\toprule
Variant & \shortstack{Variable\\chunks} & \shortstack{Mixed\\spans} & SF & Success (\%) $\uparrow$ \\
\midrule
Baseline & -- & -- & -- & $46.58\!\pm\!1.81$ \\
Baseline + 75-step span & -- & -- & -- & $40.92\!\pm\!0.80$ \\
Baseline + mixed spans & -- & $\checkmark$ & -- & $54.42\!\pm\!0.38$ \\
\midrule
New architecture & -- & -- & -- & $46.08\!\pm\!2.43$ \\
Fixed chunks + SF & -- & -- & $\checkmark$ & $48.42\!\pm\!0.38$ \\
Variable chunks & $\checkmark$ & -- & -- & $50.83\!\pm\!1.38$ \\
Variable chunks + SF & $\checkmark$ & -- & $\checkmark$ & $52.00\!\pm\!0.50$ \\
\midrule
Mixed spans, without SF & $\checkmark$ & $\checkmark$ & -- & $54.42\!\pm\!0.95$ \\
Full method & $\checkmark$ & $\checkmark$ & $\checkmark$ & $\mathbf{60.17\!\pm\!1.04}$ \\
\bottomrule
\end{tabular*}
\par\endgroup
\end{table}

\textbf{Overall performance.}
FlexiWorld achieves the highest mean success among the compared methods
across four tasks and four goal distances (Table~\ref{tab:main-results}).
Using each method's strongest evaluated configuration, FlexiWorld with
ARCEM reaches $89.29\%$ mean success, compared with $83.98\%$ for
INTACT with Guarded-A.
FlexiWorld leads on PushT, Cube, and Reacher, with the largest gains
over INTACT on the two manipulation tasks, PushT and Cube.

\textbf{Direct control and planning refinement.}
FlexiWorld's advantage is already present without search, and ARCEM
further improves its plans (Table~\ref{tab:planner-comparison}).
FlexiWorld Direct reaches $86.79\%$ mean success, compared with
$81.62\%$ for INTACT Direct, and also exceeds INTACT with Guarded-A.
With the same FlexiWorld checkpoints and observation budget, ARCEM raises
the mean success rate from $86.79\%$ to $89.29\%$ and achieves higher mean success
than Guarded-A. Its largest gain over Direct is on PushT, from
$60.39\%$ to $68.89\%$.
ARCEM also achieves higher mean success than Guarded-A at comparable
measured planning time, with Guarded-A allocated a larger search budget
(Appendix~\ref{app:search-timing}).

\textbf{Distant-goal control.}
Figure~\ref{fig:distance-results} shows stronger Direct control with
FlexiWorld on PushT and Cube at longer goal distances.
While Direct success is similar at $D=25$, FlexiWorld outperforms
INTACT on both tasks at each of the three longer distances.
At $D=100$, success increases from $12.67\%$ to $33.44\%$ on PushT
and from $84.67\%$ to $93.78\%$ on Cube.
Reacher and TwoRoom show smaller changes near saturation.
The 100-step distance exceeds the longest training span of 75 steps,
testing composition beyond the training windows.
Appendix~\ref{app:full-results} reports the complete per-distance results.

\subsection{Ablation Studies}
\label{sec:ablations}

Variable-length chunks and Student Forcing improve PushT Direct control
beyond replacing the action architecture alone (Table~\ref{tab:ablations}).
With fixed five-action chunks and SF disabled, the new action encoder
and autoregressive actor reach $46.08\%$, compared with $46.58\%$ for
the INTACT baseline. Within the new architecture, four single-span
variants form a $2\times2$ comparison at a $35$-step span and six epochs.
Variable chunks raise success from $46.08\%$ to $50.83\%$ without SF
and from $48.42\%$ to $52.00\%$ with SF. Conversely, SF improves the success
rate in both chunking settings.
On PushT at $D=25$, variable chunks without SF reach $88.67\%$, still
below the INTACT baseline's $90.33\%$; adding SF raises success to $92.67\%$.
This motivates addressing the
training--execution prefix mismatch rather than relying on variable
chunks alone: SF exposes the actor to its own generated prefixes
while retaining expert actions as targets.

Mixed-span training benefits both architectures, with the full recipe
achieving the highest success. With fixed chunks, INTACT reaches $54.42\%$
with mixed spans, versus $46.58\%$ for a 35-step span and $40.92\%$ for a
75-step span. The benefit is therefore not explained by extending the
training span alone. With the new architecture, variable chunks, and SF,
mixed spans raise success from $52.00\%$ to $60.17\%$. Removing SF reduces
success to $54.42\%$ with the same chunk and span settings.
Two mixed-span epochs and six single-span epochs provide similar totals
of supervised chunk transitions on PushT. Appendix~\ref{app:sampling}
details the schedules and additional controls.

\subsection{Planning Granularity and Further Analysis}
\label{sec:analysis}

\textbf{Longer chunks accelerate planning.}
FlexiWorld supports flexible planning granularity at deployment: the
same trained model can use different action chunk lengths to trade
planning time against control success.
Switching from $k=5$ to $k=10$ gives an average ARCEM speedup of
approximately $1.3\times$ at $D\in\{50,100\}$ while maintaining
similar four-task mean success. Predictor calls halve, but all
primitive actions remain autoregressive. Appendix~\ref{app:chunk-planning}
reports matched protocols, task-level results, and scoring-grid controls.
With Reacher and TwoRoom success near saturation, we focus the rollout
analysis on PushT and Cube. Within each checkpoint, we compare endpoint
latent prediction MSE under identical expert actions: $k=10$ reduces
this error relative to $k=5$ at all four tested distances
(Appendix~\ref{app:chunk-rollout}).

\textbf{Action prediction improves over longer horizons.}
PushT diagnostics show improved long-span action generation alongside
gains in frozen-feature probes but comparable actor-free control.
To evaluate control without learned action generation, we disable each
actor and use the same CEM planner to search actions through its world
model, obtaining $40.75\%$ success for FlexiWorld and $40.17\%$ for INTACT.
To assess information in the representations, independent ridge probes
use frozen current and goal visual features to predict either the recorded
temporal gap or the next five expert actions. Temporal-gap $R^2$ rises
from $0.565$ to $0.603$; action-probe $R^2$ is equal or higher at all
tested distances (Appendix~\ref{app:diagnostics}). These probes do not
use the learned actors. We separately evaluate the actors' next-five-action
predictions against expert actions, using matched observations, goals,
and five-action histories. FlexiWorld improves action accuracy and structural
correspondence, measured by MAE, $R^2$, nearest-neighbor overlap, and CKA,
at $D\in\{50,75,100\}$ (Figure~\ref{fig:action-diagnostics}). INTACT's
advantage at $D=25$ suggests a better fit to short goal spans, but
FlexiWorld maintains comparable Direct success at this distance while
predicting actions better over longer spans.

\textbf{Limitations and future work.}
Reliable plan selection and execution remain challenges. On PushT,
ARCEM recovers some Direct failures but also loses some Direct successes:
retaining the Direct plan among candidates does not guarantee its
selection, since ranking uses predicted latent costs rather than actual
outcomes (Appendix~\ref{app:failure-modes}). Likewise, SF exposes the
actor to generated action prefixes while retaining recorded states and
expert targets, leaving recovery from perturbed physical states untested.
Future work could investigate state perturbations, training on predicted
contexts, and uncertainty-triggered reobservation. The latter adapts
physical feedback timing rather than imagined chunk length and should
be evaluated against fixed schedules in both success and observation
and planning costs
(Appendix~\ref{app:planning-limits}).

\section{Conclusion}
FlexiWorld jointly learns latent dynamics and goal-conditioned action generation
across time scales using variable-length action chunks. Across four benchmarks,
it enables effective search-free control, and ARCEM further improves success
through actor-residual search. Longer chunks reduce planning time with
comparable mean success. A single model thus supports distant-goal control and
adjustable planning granularity without retraining, balancing planning cost and
control performance. Future work will investigate reliable candidate ranking
and evaluate FlexiWorld on physical systems.

\clearpage
\appendix
\section*{Supplementary Material}
\begingroup
\newcommand{\appendixentry}[1]{%
  \noindent\hyperref[#1]{\ref*{#1}\quad\nameref*{#1}}\dotfill
  \pageref{#1}\par}
\newcommand{\appendixsubentry}[1]{\hspace*{1em}\appendixentry{#1}}
\appendixentry{app:training}
\appendixsubentry{app:architecture}
\appendixsubentry{app:sampling}
\appendixsubentry{app:baseline-training}
\appendixentry{app:full-results}
\appendixentry{app:search}
\appendixsubentry{app:chunk-planning}
\appendixsubentry{app:search-timing}
\appendixsubentry{app:temperature}
\appendixentry{app:diagnostics}
\appendixsubentry{app:action-predictions}
\appendixsubentry{app:frozen-probes}
\appendixsubentry{app:chunk-rollout}
\appendixentry{app:failure-modes}
\appendixsubentry{app:paired-recovery}
\appendixsubentry{app:tworoom-failure}
\appendixsubentry{app:planning-limits}
\appendixsubentry{app:qualitative-rollouts}
\endgroup

\section{Implementation and Evaluation Details}
\label{app:training}

\subsection{Architecture and Training}
\label{app:architecture}
\textbf{Shared visual backbone and predictor.}
The visual encoder $\encoder$ is a randomly initialized ViT-Tiny
\citep{dosovitskiy2020vit} with $14\times14$ patches and $224\times224$
images. Visual and action embeddings have 192 dimensions, and the
projectors use hidden width 2048 with batch normalization.
The predictor $\predictor$ follows LeWM and INTACT
\citep{maes2026leworldmodel,sun2026intact}, using six causal Transformer
layers with model width 192, 16 attention heads of dimension 64,
and feed-forward width 2048. It uses dropout 0.1 and learned positional
embeddings over at most three latent states.
Actions condition each predictor layer through adaptive normalization
and residual gates.

\textbf{Variable-length action encoder.}
The causal Transformer $\actionencoder$ replaces the fixed-width action
encoder with two layers of width 64, four attention heads, and feed-forward
width 256. It projects the last valid action token to 192 dimensions and
adds a learned MLP projection of a sinusoidal chunk-length code, yielding one predictor
input regardless of the number of primitives in the chunk. The same
encoder represents the preceding action chunk for actor conditioning.

\textbf{Autoregressive actor.}
The causal Transformer $\actornet$ uses three layers of width 192, four
attention heads, and feed-forward width 768. Four conditioning tokens
encode the current latent state, intent, their elementwise product, and
the previous action chunk. The actor predicts one primitive action at
a time from this context and the within-chunk prefix, rather than
producing a fixed-width chunk in parallel. Its Gaussian output heads
predict a mean and log standard deviation, with the latter clamped to
$[-5,2]$.

\textbf{Optimization.}
FlexiWorld uses AdamW \citep{loshchilov2017adamw} with learning rate $3\times10^{-4}$, weight decay
$10^{-3}$, global batch size 256, gradient clipping at norm 1, and bfloat16
precision, with a linear-warmup cosine-annealing learning-rate scheduler.
The prediction loss has unit weight and the regularization, local
action, and goal action losses have weights $0.02$, $0.10$, and $0.05$.
SIGReg~\citep{balestriero2025lejepa,maes2026leworldmodel} uses 1024
projections and 17 quadrature knots. Fixed-chunk INTACT
and the long-window control use learning rate $5\times10^{-4}$.
Actions use training-data normalization statistics. The complete model
uses no pre-trained visual backbone, target exponential moving average (EMA),
or auxiliary temporal loss.

\textbf{Actor conditioning.}
The requested chunk length sets the number of decoded primitives, not an
additional actor conditioning token. Under the same context, deterministic
conditional-mean decoding of a shorter chunk therefore produces a prefix
of a longer decode. For Student Forcing, each chunk and intent first
receives a greedy decode without gradients. A sample-level Bernoulli draw
with probability $p_{\mathrm{SF}}$ selects these generated primitives or
the expert primitives as the within-chunk conditioning sequence. The
selected sequence is shifted right, so position $j$ receives only
positions $0,\ldots,j-1$, while its target remains expert action $j$.
Latent state, intent, and the preceding-chunk embedding are held fixed between
the two passes. The supervised pass updates the actor, action encoder,
and attached visual representation without backpropagating through
student prefix generation. The local intent retains gradients through
both boundary latent states, whereas the goal intent detaches only its
final-goal latent. The current latent and the independent prediction
targets remain attached.

\textbf{Loss aggregation.}
The compact objective in Equation~\ref{eq:odyssey-loss} preserves the
local and goal action weights above. For $q\in\{\mathrm{local},\mathrm{goal}\}$,
define
\[
 \mathcal L_q=\mathbb E\!\left[\frac{1}{N}
 \sum_{i=0}^{N-1}\ell_i^q\right],\qquad
 \mathcal L_{\mathrm{NLL}}=\mathcal L_{\mathrm{local}}
 +0.5\mathcal L_{\mathrm{goal}}.
\]
With $\lambda_{\mathrm{act}}=0.10$, the effective weights are therefore
$0.10$ for local intent and $0.05$ for goal intent. The prediction term is
\[
 \mathcal L_{\mathrm{pred}}=\mathbb E\!\left[\frac{1}{N}
 \sum_{i=0}^{N-1}
 \frac{\|\hat{\latent}_{i+1}-\latent_{i+1}\|_2^2}{d_z}\right].
\]
Here $d_z=192$ is the latent dimensionality.
The expectations cover sampled spans, windows, partitions, and prefix
choices, and are estimated by mini-batch averages. Equation~\ref{eq:chunk-action-loss}
normalizes each chunk's action NLL by its valid length $k_i$ and action
dimension $d_a$, excluding padding from both the sum and its denominator.
SIGReg is evaluated on boundary latent batches and averaged over boundaries,
with $\lambda_{\mathrm{reg}}=0.02$. Prediction gradients reach both the
predictor and target encoder branches. These definitions retain the
original per-chunk weighting rather than weighting longer chunks more heavily.

We retain LeWM's SIGReg objective to prevent representation collapse,
while SF changes only the actor's conditioning inputs.

\subsection{Sampling and Training Schedules}
\label{app:sampling}
Using a recorded future observation as a goal connects our sampling to
hindsight goal relabeling \citep{andrychowicz2017her} and supervised
goal-reaching \citep{ghosh2019gcsl}. Unlike online trajectory collection
in GCSL, our training reuses a fixed offline dataset while varying
the supervision span and intermediate chunk boundaries.

Batches use a common training span of 35, 55, or 75 primitive steps,
partitioned into $N=7$, 11, or 15 chunks, respectively.
Span groups are mixed in proportion to their available training windows.
Within each group, an episode is sampled in proportion to its valid
starting points, followed by a starting point and a bounded action partition.
Chunk lengths range from one to ten primitives. The partition preserves
the selected span and excludes the all-five schedule. Sampling is with
replacement, so an epoch specifies the number
of sampled windows. Valid-position masking excludes padding from losses.
The first chunk uses available expert history, while subsequent chunks use
the preceding chunk at its actual length.

We use two mixed-span epochs and six single-span epochs to balance
training exposure across the span groups. A mixed epoch samples from all
three span groups, whereas a single-span epoch samples from only one.
The schedules therefore allocate two passes to each of three groups or
six passes to one group. Because the groups contain different numbers of
valid windows and chunks per window, this is not an exact match in
optimizer updates or compute. On PushT, it gives similar total numbers
of supervised chunk transitions.
All variants in Tables~\ref{tab:ablations} and~\ref{tab:ablations-extended}
use training seed 0.
The no-SF mixed-span variant and Full method share the same loader and optimization settings,
changing only $p_{\mathrm{SF}}$ from zero to 0.5.
The four single-span autoregressive variants form a $2\times2$ comparison
of fixed/variable chunks and $p_{\mathrm{SF}}\in\{0,0.5\}$, using the same
action architecture, $35$-step span, and six-epoch budget.
New architecture uses seven fixed five-step chunks with SF disabled,
whereas Variable chunks partitions the same span into seven variable-length
chunks. Their SF counterparts change only the prefix-training policy.
All four results come from a unified re-evaluation with inference chunk
length five and 100 episodes per distance and evaluation seed.
Each result averages evaluation seeds 0, 1, and 42.
No action feedback retains the new encoder and actor parameterization,
but replaces within-chunk action-conditioning values with zeros during
training and inference. It retains preceding-chunk context and predicts
the current chunk in parallel, using variable partitions and no SF.
In its independent matched re-evaluation, removing within-chunk action
feedback reduces the success rate from $50.92\%$ to $48.83\%$
(Table~\ref{tab:ablations-extended}). This control
tests action feedback within this architecture, not autoregressive versus
parallel actors in general.

\begin{table}[!htb]
\centering\setlength{\tabcolsep}{3pt}
\caption{\textbf{Extended PushT component study.} Success (\%) averages
four goal distances with one replan. Sample SD is across three evaluation
seeds; training settings follow Appendix~\ref{app:sampling}. Goal spans are in
primitive steps. AR denotes autoregressive generation; SF denotes Student
Forcing, with generated-prefix probability $p_{\mathrm{SF}}$.}
\label{tab:ablations-extended}
\begingroup\small
\begin{tabular*}{\linewidth}{@{\extracolsep{\fill}}lclccr@{}}
\toprule
Variant & Goal span & Action interface & $p_{\mathrm{SF}}$ & Epochs & Success (\%) \\
\midrule
New architecture & 35 & Fixed, AR & 0 & 6 & $46.08\!\pm\!2.43$ \\
Fixed chunks + SF & 35 & Fixed, AR & 0.5 & 6 & $48.42\!\pm\!0.38$ \\
Variable chunks & 35 & Variable, AR & 0 & 6 & $50.83\!\pm\!1.38$ \\
Variable chunks + SF & 35 & Variable, AR & 0.5 & 6 & $52.00\!\pm\!0.50$ \\
No action feedback & 35 & Variable, parallel & 0 & 6 & $48.83\!\pm\!1.94$ \\
INTACT & 35 & Fixed & -- & 6 & $46.58\!\pm\!1.81$ \\
Long span & 75 & Fixed & -- & 6 & $40.92\!\pm\!0.80$ \\
Mixed spans (INTACT) & 35/55/75 & Fixed & -- & 2 & $54.42\!\pm\!0.38$ \\
Without SF & 35/55/75 & Variable, AR & 0 & 2 & $54.42\!\pm\!0.95$ \\
Full method & 35/55/75 & Variable, AR & 0.5 & 2 & $60.17\!\pm\!1.04$ \\
\bottomrule
\end{tabular*}
\par\smallskip
\begin{minipage}{\linewidth}
No action feedback ($48.83\%$) is compared with its independently
re-evaluated, matched autoregressive reference ($50.92\%$). The
$50.83\%$ Variable chunks result above is
from the factorial re-evaluation of the same reference checkpoint, not
the matched reference for this comparison.
\end{minipage}
\par\endgroup
\end{table}

The additional fixed-interface control distinguishes a single long span
from mixed-span supervision: training only at 75 steps reaches $40.92\%$,
compared with $54.42\%$ for mixed spans
(Table~\ref{tab:ablations-extended}). The remaining component comparisons
are discussed in Section~\ref{sec:ablations}.

\subsection{Baselines and Evaluation Protocol}
\label{app:baseline-training}
\label{app:statistics}
\textbf{Baseline models and training budgets.}
For all baselines, we use the configurations provided by their official
implementations, with training-budget and evaluation adaptations specified below.
We evaluate LeWM, Fast-LeWM, Sub-JEPA, and PushT DINO-WM with their
available models, retaining their respective training budgets. DINO-WM
also uses a pre-trained DINOv2 encoder \citep{zhou2024dinowm,oquab2023dinov2},
whereas FlexiWorld trains from scratch. Our six-epoch INTACT control
provides a comparison at a similar supervision budget, detailed
in Appendix~\ref{app:sampling}; the external baselines are not retrained
under a common compute budget. Baseline success rates are from our own
evaluations under the one-replan protocol below, without substituting
published success rates.

\textbf{Evaluation protocol.}
Goal observations are recorded at $t+D$, never rounded to a multiple of five.
The main evaluation allows up to $2D$ primitive actions, with one new
observation and replan after the first $D$ actions if needed. Tables label
this setting ``One replan''. ``Strict'' uses only the initial $D$-step
open-loop plan, without a new observation or replan.
Success uses the environment predicate and is latched within the budget.
On PushT the configured agent-position condition is included. Models
receive images rather than privileged state variables, except the released
PushT DINO-WM checkpoint, which also uses proprioception. All methods receive
the same goal observation and execution budget.
The controlled search and timing experiments use the math backend of
scaled dot-product attention (Math SDPA) and an evaluation batch size of one.

\textbf{Seeds and aggregation.}
FlexiWorld uses training seeds $\{0,42,3072\}$ and evaluation seeds
$\{0,1,42\}$, with 100 episodes per distance and evaluation seed.
Baseline entries use one training seed and the same three evaluation seeds.
For each FlexiWorld training seed, success is averaged over evaluation seeds
and the four distances. Task entries report the mean and sample SD of
these three training-seed means. For Average, we first average the four tasks within
each training seed, then compute the mean and sample SD. For one-checkpoint
controls, we instead report SD across evaluation-seed means.
Unless otherwise stated, subsequent control evaluations follow these seed
settings and aggregation rules.

DINO-WM uses the author-released checkpoint on PushT and our ten-epoch,
pixels-only checkpoints at training seed 0 on Cube, Reacher, and TwoRoom.
These task-specific checkpoints follow the one-checkpoint aggregation above;
for Average, task means are averaged within each evaluation seed before
computing the mean and sample SD.

Run-to-run variability and aggregation choices can affect empirical
comparisons \citep{henderson2017deeprl,agarwal2021evaluation}.
The reported SD describes variability across the specified repeats,
not a confidence interval or a test of statistical significance.

\section{FlexiWorld Results by Goal Distance}
\label{app:full-results}
Table~\ref{tab:full-results} expands the main results by goal distance for
Direct and ARCEM.

\begin{table}[htbp]
\centering\setlength{\tabcolsep}{3pt}
\caption{\textbf{FlexiWorld success by goal distance.} Success (\%) allows
one replan. Entries show mean and SD across three training seeds after
averaging evaluation seeds. $D$ is measured in primitive steps. The last
column averages distances within each training seed before computing SD.}
\label{tab:full-results}
\begingroup\small
\begin{tabular*}{\linewidth}{@{\extracolsep{\fill}}llrrrrr@{}}
\toprule
Planner & Task & $D=25$ & $D=50$ & $D=75$ & $D=100$ & Mean \\
\midrule
Direct & PushT & $89.67\!\pm\!1.67$ & $71.56\!\pm\!3.98$ & $46.89\!\pm\!5.36$ & $33.44\!\pm\!5.35$ & $60.39\!\pm\!3.92$ \\
Direct & Cube & $100.00\!\pm\!0.00$ & $84.67\!\pm\!2.40$ & $87.00\!\pm\!3.06$ & $93.78\!\pm\!0.84$ & $91.36\!\pm\!1.56$ \\
Direct & Reacher & $98.11\!\pm\!0.19$ & $99.89\!\pm\!0.19$ & $99.56\!\pm\!0.19$ & $98.67\!\pm\!0.33$ & $99.06\!\pm\!0.05$ \\
Direct & TwoRoom & $100.00\!\pm\!0.00$ & $99.11\!\pm\!0.51$ & $96.44\!\pm\!3.36$ & $89.89\!\pm\!3.89$ & $96.36\!\pm\!1.92$ \\
ARCEM & PushT & $94.33\!\pm\!2.08$ & $78.11\!\pm\!4.55$ & $55.00\!\pm\!8.01$ & $48.11\!\pm\!5.35$ & $68.89\!\pm\!4.80$ \\
ARCEM & Cube & $100.00\!\pm\!0.00$ & $88.78\!\pm\!2.52$ & $86.22\!\pm\!3.02$ & $92.78\!\pm\!1.02$ & $91.94\!\pm\!1.50$ \\
ARCEM & Reacher & $99.89\!\pm\!0.19$ & $100.00\!\pm\!0.00$ & $99.89\!\pm\!0.19$ & $99.11\!\pm\!0.51$ & $99.72\!\pm\!0.17$ \\
ARCEM & TwoRoom & $99.89\!\pm\!0.19$ & $98.56\!\pm\!1.17$ & $96.22\!\pm\!3.42$ & $91.78\!\pm\!5.09$ & $96.61\!\pm\!2.45$ \\
\bottomrule
\end{tabular*}
\par\endgroup
\end{table}

\section{Planning Granularity, Planning Time, and Sensitivity}
\label{app:search}
\subsection{Planning with Five- and Ten-Action Chunks}
\label{app:chunk-planning}

\textbf{Matched evaluation.}
We vary deployment chunk length without changing the learned weights.
Direct and ARCEM are evaluated at $k=5$ and $k=10$ on all four tasks and
$D\in\{25,50,75,100\}$, with identical starts and goals across chunk
schedules. Both schedules generate exactly
$D$ primitive actions per plan, with $k=10$ using a final five-action chunk at
$D=25$ and $D=75$. Both permit one reobservation after the first plan.
The initial expert history and subsequent executed history each contain
five primitive actions, independently of the planned chunk length.
These matched re-evaluations are reported separately from Table~\ref{tab:main-results}.

ARCEM uses additive residuals with no actor-standard-deviation scaling,
$T=0.2$, 128 candidates, three
iterations, and 16 elites for both schedules. Candidate costs use the
predicted chunk-boundary states. To distinguish rollout granularity from
the set of scoring times, a third ARCEM control retains $k=5$ rollouts
but scores only the boundary times of the $k=10$ schedule, including
the final endpoint. Direct has 288 evaluation combinations and ARCEM has 432,
including this common-grid control. Table~\ref{tab:chunk-success}
summarizes success by task, Table~\ref{tab:chunk-latency} compares success
and planning time by distance, and Table~\ref{tab:chunk-distances} provides
the complete task-by-distance results.

\begin{table}[!htbp]
\centering\setlength{\tabcolsep}{3pt}
\caption{\textbf{Planning chunk length and control success.} $k$ is the chunk length in primitive actions. Success (\%) with one replan averages four goal distances. Mean and sample SD across three training seeds, after averaging evaluation seeds and distances. All rows use the matched chunk-length study; they do not replace the main-result evaluation. $\dagger$: common scoring grid, retaining five-action rollouts but scoring only the ten-action boundary times.}
\label{tab:chunk-success}
\begingroup\small
\begin{tabular*}{\linewidth}{@{\extracolsep{\fill}}llrrrrr@{}}
\toprule
Planner & $k$ & PushT & Cube & Reacher & TwoRoom & Average \\
\midrule
Direct & 5 & $60.39\!\pm\!3.92$ & $91.36\!\pm\!1.56$ & $99.06\!\pm\!0.05$ & $96.36\!\pm\!1.92$ & $86.79\!\pm\!1.65$ \\
Direct & 10 & $54.39\!\pm\!1.35$ & $91.50\!\pm\!1.32$ & $99.25\!\pm\!0.25$ & $96.22\!\pm\!1.88$ & $85.34\!\pm\!0.92$ \\
ARCEM & 5 & $68.89\!\pm\!4.80$ & $91.94\!\pm\!1.50$ & $99.72\!\pm\!0.17$ & $96.61\!\pm\!2.45$ & $89.29\!\pm\!1.96$ \\
ARCEM & 10 & $67.14\!\pm\!1.82$ & $92.11\!\pm\!1.93$ & $99.69\!\pm\!0.05$ & $97.72\!\pm\!1.42$ & $89.17\!\pm\!1.08$ \\
ARCEM & 5$^\dagger$ & $68.50\!\pm\!4.13$ & $91.72\!\pm\!1.88$ & $99.69\!\pm\!0.17$ & $96.83\!\pm\!2.17$ & $89.19\!\pm\!1.85$ \\
\bottomrule
\end{tabular*}
\par\endgroup
\end{table}

\textbf{Control trade-off.}
Increasing chunk length leaves ARCEM's four-task mean close to its
$k=5$ value ($89.17\%$ versus $89.29\%$), but the changes are not
uniform. PushT decreases from $68.89\%$ to $67.14\%$, while TwoRoom
increases from $96.61\%$ to $97.72\%$. Direct is more sensitive on
PushT, decreasing from $60.39\%$ to $54.39\%$, while its other task means
change little. The common-grid ARCEM control reaches $89.19\%$
overall, with $68.50\%$ on PushT. Thus ARCEM maintains similar average
success with longer chunks, while individual tasks and Direct control
remain sensitive to the chunk schedule.

\begin{table}[!htbp]
\centering\setlength{\tabcolsep}{3pt}
\caption{\textbf{Matched success and planning time by goal distance.} SR denotes success rate (\%), averaged over four tasks with training-seed SD. $D$ is goal distance and $k$ is chunk length, both in primitive steps. Planning time averages per-input median solve times on identical inputs, including encoding but excluding environment execution. Speedup is the ratio of mean planning times, not an episode-runtime ratio.}
\label{tab:chunk-latency}
\begingroup\small
\begin{tabular*}{\linewidth}{@{\extracolsep{\fill}}llrrrrr@{}}
\toprule
Planner & $D$ & SR, $k=5$ & SR, $k=10$ & ms, $k=5$ & ms, $k=10$ & Speedup \\
\midrule
Direct & 25 & $96.94\!\pm\!0.38$ & $96.86\!\pm\!0.25$ & 64.2 & 55.1 & 1.17$\times$ \\
Direct & 50 & $88.81\!\pm\!1.18$ & $87.28\!\pm\!0.84$ & 115.6 & 92.8 & 1.25$\times$ \\
Direct & 75 & $82.47\!\pm\!2.70$ & $81.00\!\pm\!1.30$ & 167.2 & 135.1 & 1.24$\times$ \\
Direct & 100 & $78.94\!\pm\!2.50$ & $76.22\!\pm\!1.92$ & 218.4 & 172.6 & 1.27$\times$ \\
ARCEM & 25 & $98.53\!\pm\!0.55$ & $98.86\!\pm\!0.38$ & 268.8 & 221.5 & 1.21$\times$ \\
ARCEM & 50 & $91.36\!\pm\!1.99$ & $90.75\!\pm\!0.58$ & 510.9 & 393.6 & 1.30$\times$ \\
ARCEM & 75 & $84.33\!\pm\!3.06$ & $85.11\!\pm\!1.39$ & 753.3 & 588.1 & 1.28$\times$ \\
ARCEM & 100 & $82.94\!\pm\!2.51$ & $81.94\!\pm\!2.09$ & 994.4 & 760.1 & 1.31$\times$ \\
\bottomrule
\end{tabular*}
\par\endgroup
\end{table}

\begin{table}[!htb]
\centering\setlength{\tabcolsep}{3pt}
\caption{\textbf{Chunk-length success by task and goal distance.} Matched Direct and ARCEM evaluations with one replan. $D$ is goal distance and $k$ is chunk length, both in primitive steps. Entries give mean success (\%) and sample SD across three training seeds after averaging evaluation seeds.}
\label{tab:chunk-distances}
\begingroup\small
\begin{tabular*}{\linewidth}{@{\extracolsep{\fill}}llrrrr@{}}
\toprule
Task & $D$ & Direct, $k=5$ & Direct, $k=10$ & ARCEM, $k=5$ & ARCEM, $k=10$ \\
\midrule
PushT & 25 & $89.67\!\pm\!1.67$ & $88.00\!\pm\!1.20$ & $94.33\!\pm\!2.08$ & $95.44\!\pm\!1.50$ \\
PushT & 50 & $71.56\!\pm\!3.98$ & $64.56\!\pm\!1.39$ & $78.11\!\pm\!4.55$ & $74.44\!\pm\!2.83$ \\
PushT & 75 & $46.89\!\pm\!5.36$ & $41.22\!\pm\!0.69$ & $55.00\!\pm\!8.01$ & $56.44\!\pm\!1.50$ \\
PushT & 100 & $33.44\!\pm\!5.35$ & $23.78\!\pm\!3.91$ & $48.11\!\pm\!5.35$ & $42.22\!\pm\!5.42$ \\
Cube & 25 & $100.00\!\pm\!0.00$ & $100.00\!\pm\!0.00$ & $100.00\!\pm\!0.00$ & $100.00\!\pm\!0.00$ \\
Cube & 50 & $84.67\!\pm\!2.40$ & $85.67\!\pm\!2.40$ & $88.78\!\pm\!2.52$ & $89.22\!\pm\!4.60$ \\
Cube & 75 & $87.00\!\pm\!3.06$ & $86.67\!\pm\!2.33$ & $86.22\!\pm\!3.02$ & $86.67\!\pm\!3.71$ \\
Cube & 100 & $93.78\!\pm\!0.84$ & $93.67\!\pm\!0.58$ & $92.78\!\pm\!1.02$ & $92.56\!\pm\!0.38$ \\
Reacher & 25 & $98.11\!\pm\!0.19$ & $99.44\!\pm\!0.19$ & $99.89\!\pm\!0.19$ & $100.00\!\pm\!0.00$ \\
Reacher & 50 & $99.89\!\pm\!0.19$ & $99.78\!\pm\!0.38$ & $100.00\!\pm\!0.00$ & $99.89\!\pm\!0.19$ \\
Reacher & 75 & $99.56\!\pm\!0.19$ & $99.56\!\pm\!0.51$ & $99.89\!\pm\!0.19$ & $99.78\!\pm\!0.38$ \\
Reacher & 100 & $98.67\!\pm\!0.33$ & $98.22\!\pm\!0.19$ & $99.11\!\pm\!0.51$ & $99.11\!\pm\!0.51$ \\
TwoRoom & 25 & $100.00\!\pm\!0.00$ & $100.00\!\pm\!0.00$ & $99.89\!\pm\!0.19$ & $100.00\!\pm\!0.00$ \\
TwoRoom & 50 & $99.11\!\pm\!0.51$ & $99.11\!\pm\!0.51$ & $98.56\!\pm\!1.17$ & $99.44\!\pm\!0.38$ \\
TwoRoom & 75 & $96.44\!\pm\!3.36$ & $96.56\!\pm\!3.01$ & $96.22\!\pm\!3.42$ & $97.56\!\pm\!2.01$ \\
TwoRoom & 100 & $89.89\!\pm\!3.89$ & $89.22\!\pm\!4.00$ & $91.78\!\pm\!5.09$ & $93.89\!\pm\!3.34$ \\
\bottomrule
\end{tabular*}
\par\endgroup
\end{table}

\subsection{Search Budget and Planning Time}
\label{app:search-timing}
Pure CEM uses a zero-mean normalized-action proposal with scale 1,
300 candidates, 30 iterations, and 30 elites. Guarded-A~\citep{sun2026intact} uses scale 0.25,
128 candidates, three iterations, and 16 elites, retaining the Direct
reference. ARCEM searches additive residuals in normalized action coordinates with the latter
candidate budget and $T=0.2$. Its residual distribution starts with zero
mean and unit standard deviation. Each update replaces the distribution
with the elite mean and population standard deviation, clipping the latter
to $[0.05,2.0]$, with no smoothing or residual penalty. The 128 candidates
include one slot for the exact Direct plan and one for the best residual
candidate retained across iterations. Equal candidate counts need not imply equal computational
cost because ARCEM regenerates conditional actions. Table~\ref{tab:matched-search}
uses matched checkpoints, Math SDPA, batch size, and evaluation cases for success.

\begin{figure}[!htbp]
\centering
\includegraphics[width=\linewidth]{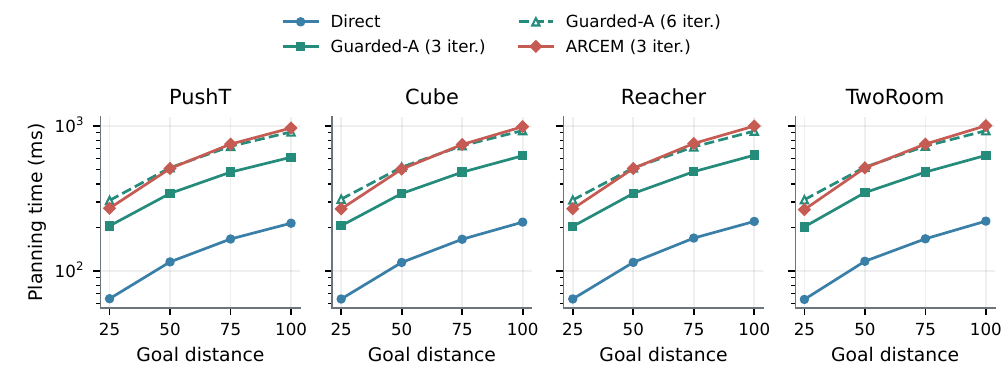}
\caption{\textbf{Planning time by goal distance.} FlexiWorld with Direct,
ARCEM (three iterations), and Guarded-A (three or six iterations).
Curves average per-input median times over both solve stages, including
image encoding and planning. The six-iteration run uses the same GPU and
input cases. The vertical axis shows per-solve time on a logarithmic scale.}
\label{fig:runtime-distance}
\end{figure}

At the same budget of 384 candidates, ARCEM
outperforms Guarded-A on PushT, Cube, and Reacher, but takes longer per solve.
Increasing Guarded-A to six iterations raises its average success from
87.02\% to 87.44\%, compared with 88.96\% for ARCEM.
On the matched input cases, six-iteration Guarded-A takes 620.5 ms
per solve and ARCEM takes 631.8 ms, placing the two within 1.8\% in mean
planning time.
Figure~\ref{fig:runtime-distance} shows planning time by distance for
Direct, ARCEM, and both the three- and six-iteration Guarded-A variants.

\begin{table}[!htb]
\centering\setlength{\tabcolsep}{3pt}
\caption{\textbf{Planning success and time with FlexiWorld.}
Success (\%) averages four goal distances with one replan; Average also
averages four tasks. Success evaluations share the seed-0 checkpoints, the math backend of scaled dot-product
attention (Math SDPA), and batch size one; mean and SD are across the three evaluation seeds.
Candidates gives the total search budget per solve across all iterations.
Planning time uses identical inputs on one RTX PRO 6000.
$\dagger$ denotes six iterations of 128 candidates.}
\label{tab:matched-search}
\begingroup\small
\begin{tabular*}{\linewidth}{@{\extracolsep{\fill}}lcrrrrrr@{}}
\toprule
Planner & Candidates & PushT & Cube & Reacher & TwoRoom & Average & Time (ms) \\
\midrule
Direct & 0 & $60.17\!\pm\!1.04$ & $89.67\!\pm\!1.84$ & $99.08\!\pm\!0.52$ & $96.42\!\pm\!1.01$ & $86.33\!\pm\!0.55$ & 141.3 \\
Guarded-A & 384 & $63.50\!\pm\!2.14$ & $88.67\!\pm\!2.27$ & $98.08\!\pm\!1.01$ & $97.83\!\pm\!0.38$ & $87.02\!\pm\!0.49$ & 414.1 \\
Guarded-A$^\dagger$ & 768 & $65.83\!\pm\!2.36$ & $88.75\!\pm\!1.56$ & $97.17\!\pm\!1.04$ & $98.00\!\pm\!0.50$ & $87.44\!\pm\!0.76$ & 620.5 \\
ARCEM & 384 & $69.67\!\pm\!2.38$ & $90.42\!\pm\!2.04$ & $99.58\!\pm\!0.38$ & $96.17\!\pm\!1.38$ & $88.96\!\pm\!0.67$ & 631.8 \\
\bottomrule
\end{tabular*}
\par\endgroup
\end{table}

Planning time is measured with
PyTorch 2.7.1 \citep{paszke2019pytorch}, CUDA 12.8, Math SDPA, and batch size one
on one RTX PRO 6000.
Each shared input case uses three warm-up solves
and 20 timed solves with CUDA synchronization. Timing covers encoding and
planning, excluding data loading, environment execution, and logging.
Second-stage inputs come from a common reference policy. Measurements use
identical initial and reobservation input cases, with one case per task and
distance at each stage. We take the median of the 20 timed solves for each
case, then average these medians over tasks, distances, and both stages.
These fixed-input planning-time measurements are separate from episode-level
success evaluations.

\FloatBarrier
\subsection{Temperature Sensitivity}
\label{app:temperature}

\begin{figure}[!htbp]
\centering
\includegraphics[width=\linewidth]{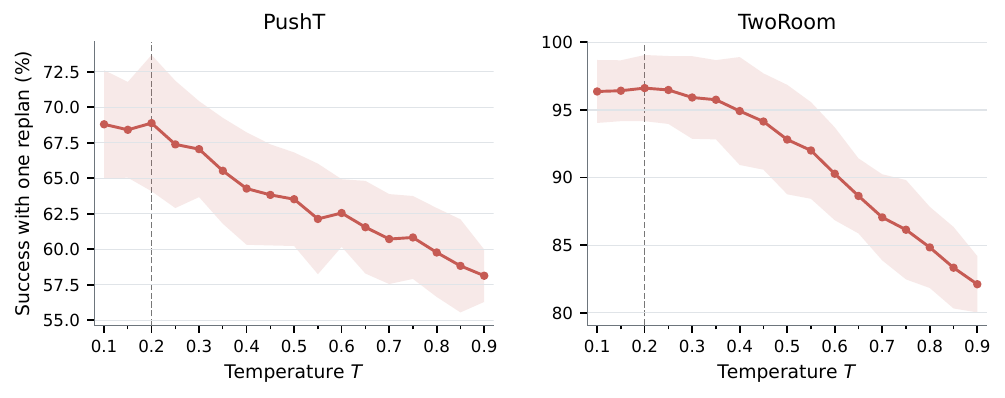}
\caption{\textbf{ARCEM temperature sensitivity at 0.05 resolution.}
Markers show measured settings from $T=0.10$ to $0.90$, with lines connecting
the means and bands showing training-seed SD after averaging evaluation seeds
and goal distances. Dashed lines mark the main setting $T=0.2$.
Both panels use the main candidate-scoring rule and one replan.}
\label{fig:temperature-sensitivity}
\end{figure}

We evaluate $T=0.10,0.15,\ldots,0.90$ on PushT and TwoRoom, holding
checkpoints, sampled starts and goals, $k=5$, and the search budget fixed.
Every temperature covers four goal distances with 100 episodes per cell,
for 1,224 completed cells across both tasks. Residuals are additive and are not multiplied
by the actor's predicted standard deviation. Candidate scoring follows
the main evaluation, before environment clipping.
Table~\ref{tab:temperature-pusht} lists the results, and
Figure~\ref{fig:temperature-sensitivity} shows the temperature trends.

Modest perturbations preserve control success better than large search
radii. PushT scores $68.81\pm3.81\%$ at $T=0.1$ and
$68.89\pm4.80\%$ at the main setting $T=0.2$, falling to
$58.14\pm1.85\%$ at $T=0.9$. TwoRoom follows the same broad pattern,
with $96.36\pm2.33\%$, $96.61\pm2.45\%$, and $82.11\pm2.08\%$
at these temperatures. The differences between adjacent settings do not
establish a sharp optimum, but the decline at larger temperatures
motivates examining whether perturbed commands remain executable.
Appendix~\ref{app:tworoom-failure} investigates this mismatch at $T=0.8$.

\begin{table}[!htbp]
\centering\setlength{\tabcolsep}{3pt}
\caption{\textbf{Dense ARCEM temperature sweep.} $T$ scales action residuals in normalized coordinates. Success (\%) with one replan at $T=0.10,0.15,\ldots,0.90$. Mean and sample SD across three training seeds after averaging three evaluation seeds and four goal distances; 100 episodes per cell. All 1,224 cells are complete, using additive residuals without actor-standard-deviation scaling and the main candidate-scoring rule.}
\label{tab:temperature-pusht}
\begingroup\small
\begin{tabular*}{\linewidth}{@{\extracolsep{\fill}}lrrlrr@{}}
\toprule
$T$ & PushT & TwoRoom & $T$ & PushT & TwoRoom \\
\midrule
0.10 & $68.81\!\pm\!3.81$ & $96.36\!\pm\!2.33$ & 0.55 & $62.14\!\pm\!3.89$ & $92.00\!\pm\!3.57$ \\
0.15 & $68.42\!\pm\!3.37$ & $96.42\!\pm\!2.25$ & 0.60 & $62.56\!\pm\!2.38$ & $90.28\!\pm\!3.43$ \\
0.20 & $68.89\!\pm\!4.80$ & $96.61\!\pm\!2.45$ & 0.65 & $61.56\!\pm\!3.26$ & $88.64\!\pm\!2.77$ \\
0.25 & $67.39\!\pm\!4.48$ & $96.47\!\pm\!2.51$ & 0.70 & $60.72\!\pm\!3.17$ & $87.06\!\pm\!3.18$ \\
0.30 & $67.06\!\pm\!3.38$ & $95.92\!\pm\!3.06$ & 0.75 & $60.83\!\pm\!2.92$ & $86.14\!\pm\!3.67$ \\
0.35 & $65.53\!\pm\!3.73$ & $95.75\!\pm\!2.93$ & 0.80 & $59.78\!\pm\!3.13$ & $84.83\!\pm\!3.00$ \\
0.40 & $64.28\!\pm\!3.96$ & $94.92\!\pm\!3.99$ & 0.85 & $58.83\!\pm\!3.28$ & $83.33\!\pm\!3.01$ \\
0.45 & $63.83\!\pm\!3.56$ & $94.14\!\pm\!3.56$ & 0.90 & $58.14\!\pm\!1.85$ & $82.11\!\pm\!2.08$ \\
0.50 & $63.53\!\pm\!3.30$ & $92.81\!\pm\!4.05$ &  &  &  \\
\bottomrule
\end{tabular*}
\par\endgroup
\end{table}

\section{Action and Representation Diagnostics}
\label{app:diagnostics}
\subsection{Goal-Conditioned Action Predictions}
\label{app:action-predictions}
We compare INTACT and FlexiWorld on matched PushT examples using three
independently trained checkpoints per model, with training seeds
$\{0,42,3072\}$. At $D\in\{25,50,75,100\}$, the shared evaluation set
contains 493, 483, 461, and 441 valid examples, respectively.
We compute each metric over this set for each checkpoint, then report
the mean and sample SD across training seeds. Each actor receives the encoded
current observation, goal observation, and preceding five expert actions, and predicts
the next five primitive actions. Both actors use conditional means: INTACT predicts
the five-action chunk jointly, whereas FlexiWorld generates it autoregressively
using its own predicted action prefix. This measures action prediction from encoded
observations rather than an autoregressive latent rollout. Each sequence is
flattened into normalized action coordinates. Mean absolute error (MAE) averages
absolute errors over examples and coordinates. We compute $R^2$ as one
minus the total squared error divided by the total squared deviation of
expert actions from their per-coordinate means, rather than averaging
coordinate-wise $R^2$ values. For each example, we find its ten nearest
neighbors by Euclidean distance separately in the predicted and expert
action matrices, excluding itself. Neighbor overlap is the size of the
intersection divided by ten, averaged over examples. Linear centered kernel alignment (CKA) compares
the column-centered action matrices \citep{kornblith2019cka}.
\begin{figure}[!htb]
\centering
\includegraphics[width=\linewidth]{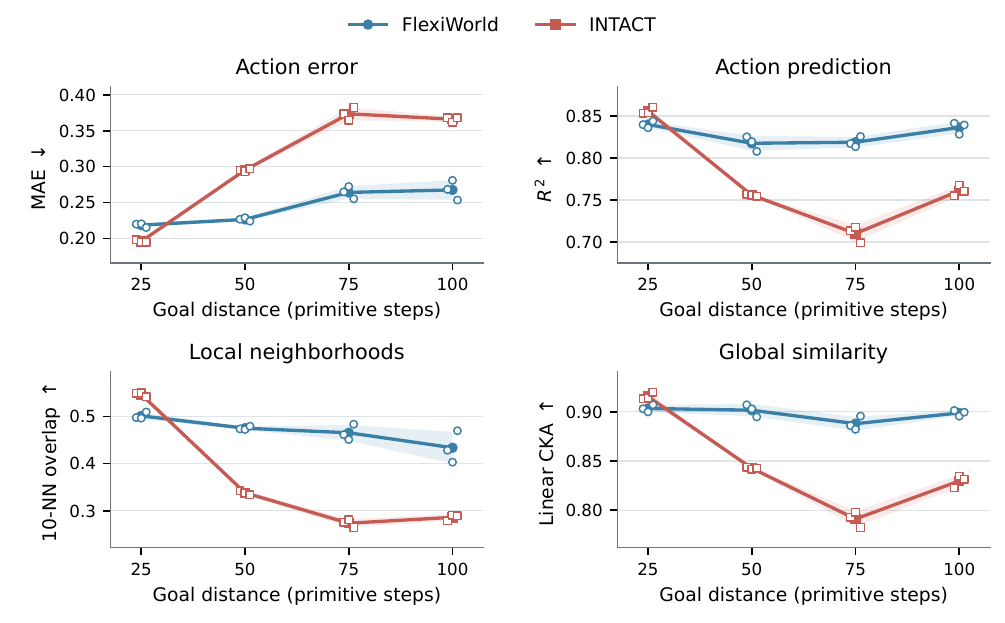}
\caption{\textbf{Goal-conditioned action correspondence on PushT.}
Lines show means and shaded bands show sample SD across three training
seeds. Hollow points show individual checkpoints, slightly offset
horizontally for visibility. Both models predict the next five expert
actions from matched observations, goals, and preceding expert chunks.
Arrows indicate the preferred direction for each metric. INTACT is better
at $D=25$, whereas FlexiWorld is better at $D\in\{50,75,100\}$ on all four mean metrics.}
\label{fig:action-diagnostics}
\end{figure}

\subsection{Frozen Probes and Actor-Free Planning}
\label{app:frozen-probes}
\begin{figure}[!htbp]
\centering
\includegraphics[width=\linewidth]{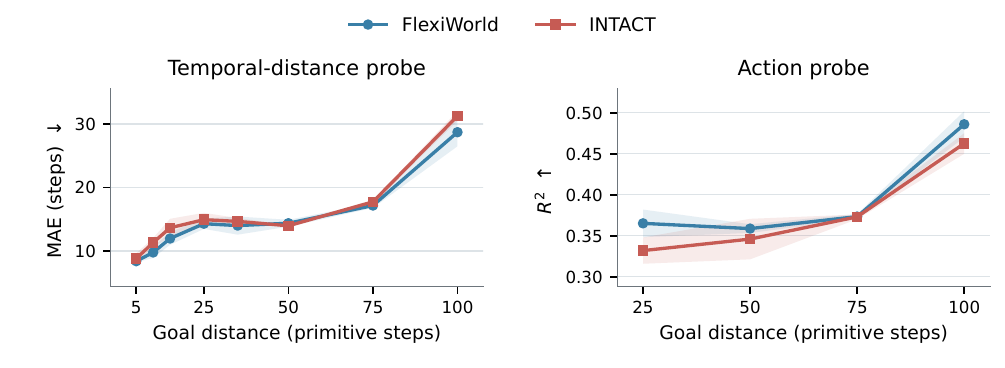}
\caption{\textbf{Frozen-feature probes on PushT.}
Left: temporal-gap MAE. Right: action $R^2$ from frozen-feature ridge probes.
Curves show means and sample SD across three training seeds,
using each diagnostic's matched sample set. The temporal probe predicts demonstrated
time gaps, not minimum time to a goal. Neither probe updates the world model.}
\label{fig:dynamics-probes}
\end{figure}

\textbf{Frozen-feature probes.}
Independent ridge regressions assess information in the visual features
without using the learned actor. Each probe is fitted separately for each
checkpoint at training seeds $\{0,42,3072\}$, using matched samples across
models and checkpoints. Reported means and sample SDs are computed across
these three checkpoint-specific fits. Both probes use
$[\latent_0,\latent_g-\latent_0,\latent_0\odot(\latent_g-\latent_0)]$, where
$\latent_0=\encoder(\obs_s)$ and $\latent_g=\encoder(\obs_g)$.
Neither probe receives action history.
The action probe predicts the next five expert primitives from frozen
visual features. It is fitted jointly on samples at
$D\in\{25,50,75,100\}$ and evaluated separately at each distance.
An episode-disjoint 80/20 split is used for probe fitting and
evaluation, not for world-model pretraining. Input coordinates are
standardized using fitting-set means and standard deviations, and the
ridge coefficient is 0.01.
A temporal-distance ridge probe uses the same frozen feature vector to
predict the recorded number of primitive steps between the current and
goal observations, using the same split procedure, standardization, and
ridge coefficient.
The probe is fitted jointly across eight distances
$5,10,15,25,35,50,75,100$ and evaluated on 256 examples per distance.
Its overall $R^2$ pools examples across these distances, while
Figure~\ref{fig:dynamics-probes} reports MAE separately at each distance.
Its $R^2$ is $0.565\pm0.026$ for INTACT and
$0.603\pm0.029$ for FlexiWorld, measuring prediction of demonstrated
time gaps rather than minimum control times.
Table~\ref{tab:representation} reports action-probe
scores and representation statistics. Effective rank is the exponential
of the entropy of the normalized singular values of the centered current-state
latent matrix, and latent standard deviation is averaged over coordinates.
Both models retain variation across examples, with lower effective rank
for FlexiWorld. The probes measure information accessible to ridge
regression, rather than representation quality for every downstream use.

\begin{table}[htbp]
\centering\setlength{\tabcolsep}{3pt}
\caption{\textbf{Frozen-feature probes and representation statistics on PushT.}
Entries report mean and SD across three training seeds on matched samples.
The action probe predicts five expert primitives. Effective rank and latent
standard deviation (std.) describe the current-state representations.
$R^2$ is the coefficient of determination; $D$ is goal distance in primitive steps.}
\label{tab:representation}
\begingroup\small
\begin{tabular*}{\linewidth}{@{\extracolsep{\fill}}lrrrr@{}}
\toprule
Model & $D$ & Probe $R^2$ & Effective rank & Latent std. \\
\midrule
INTACT & 25 & $0.332\!\pm\!0.016$ & $81.3\!\pm\!9.5$ & $0.981\!\pm\!0.015$ \\
INTACT & 50 & $0.346\!\pm\!0.025$ & $79.3\!\pm\!9.3$ & $0.980\!\pm\!0.017$ \\
INTACT & 75 & $0.373\!\pm\!0.003$ & $75.1\!\pm\!9.0$ & $0.976\!\pm\!0.015$ \\
INTACT & 100 & $0.463\!\pm\!0.013$ & $72.2\!\pm\!8.4$ & $0.990\!\pm\!0.015$ \\
FlexiWorld & 25 & $0.365\!\pm\!0.017$ & $53.0\!\pm\!12.2$ & $0.976\!\pm\!0.011$ \\
FlexiWorld & 50 & $0.359\!\pm\!0.005$ & $52.0\!\pm\!12.1$ & $0.973\!\pm\!0.013$ \\
FlexiWorld & 75 & $0.373\!\pm\!0.003$ & $49.4\!\pm\!11.5$ & $0.962\!\pm\!0.013$ \\
FlexiWorld & 100 & $0.486\!\pm\!0.016$ & $47.1\!\pm\!11.0$ & $0.972\!\pm\!0.014$ \\
\bottomrule
\end{tabular*}
\par\endgroup
\end{table}

\textbf{Actor-free control.}
We disable both learned actors and search actions using pure CEM, with
300 candidates, 30 iterations, and 30 elites. On PushT, both models use
$k=5$ at $D\in\{25,50,75,100\}$. This comparison uses one
checkpoint per model at training seed 0.
Averaging success over the evaluation repeats and four goal distances
gives $40.75\%$ for FlexiWorld and $40.17\%$ for INTACT with one replan,
compared with $35.67\%$ and $35.83\%$ after the first plan, respectively.
The two models perform similarly when neither actor is used.

\subsection{Within-Model Rollout Error across Chunk Lengths}
\label{app:chunk-rollout}

\textbf{Purpose and protocol.}
On PushT and Cube, this diagnostic compares endpoint prediction at $k=5$ and
$k=10$ within the same checkpoint under identical expert actions. We reuse the episode IDs and
starting steps from the matched Direct study in
Appendix~\ref{app:chunk-planning}. Each schedule receives exactly the
same normalized expert primitive actions and initial five-action expert
history. We roll the predictor forward without intermediate observations,
encode the recorded image at $t+D$ with the same checkpoint, and compute
mean squared error over visual-latent coordinates at that endpoint.
The $k=10$ schedule again uses a final five-action chunk when required.
The measurement covers the initial $D$-step horizon, not the later replan.

The reported comparison contains 7,200 paired examples across these two tasks,
four distances, and the three checkpoints per task from the matched Direct study.
For each task, checkpoint, and distance, we average 300 examples before
computing the mean and sample SD across checkpoints.
Table~\ref{tab:chunk-rollout} reports all distances, including $D=50$
and $D=100$ discussed in the main text.

\begin{table}[!htbp]
\centering\setlength{\tabcolsep}{3pt}
\caption{\textbf{Expert-action endpoint prediction under two chunk schedules.} Latent mean squared error (MSE) after a $D$-step rollout ($k$: chunk length in primitive actions), with identical expert primitive actions and no intermediate observations. Mean and sample SD across three training seeds; each seed averages 300 matched examples per task and distance. Changes compare $k=10$ with $k=5$ within the same model; latent errors are not pooled across tasks.}
\label{tab:chunk-rollout}
\begingroup\small
\begin{tabular*}{\linewidth}{@{\extracolsep{\fill}}llrrr@{}}
\toprule
Task & $D$ & MSE, $k=5$ & MSE, $k=10$ & Relative change \\
\midrule
PushT & 25 & $0.0408\!\pm\!0.0017$ & $0.0391\!\pm\!0.0023$ & -4.13\% \\
PushT & 50 & $0.1604\!\pm\!0.0180$ & $0.1541\!\pm\!0.0120$ & -3.90\% \\
PushT & 75 & $0.3532\!\pm\!0.0227$ & $0.3108\!\pm\!0.0358$ & -12.00\% \\
PushT & 100 & $0.5797\!\pm\!0.0243$ & $0.5297\!\pm\!0.0388$ & -8.61\% \\
Cube & 25 & $0.0315\!\pm\!0.0017$ & $0.0177\!\pm\!0.0013$ & -43.70\% \\
Cube & 50 & $0.0356\!\pm\!0.0027$ & $0.0278\!\pm\!0.0005$ & -21.93\% \\
Cube & 75 & $0.0606\!\pm\!0.0069$ & $0.0386\!\pm\!0.0019$ & -36.33\% \\
Cube & 100 & $0.0639\!\pm\!0.0061$ & $0.0404\!\pm\!0.0021$ & -36.74\% \\
\bottomrule
\end{tabular*}
\par\endgroup
\end{table}

\textbf{Effect of chunk length.}
Ten-action chunks reduce mean expert-action endpoint MSE at every tested
distance on PushT and Cube. On PushT, MSE changes from $0.1604$ to $0.1541$ at $D=50$
and from $0.5797$ to $0.5297$ at $D=100$, yet Direct success decreases
under the longer chunks. Cube combines lower MSE with similar success.
Lower error under expert actions need not improve control, which depends
on generated actions and candidate selection.

\section{Execution Outcomes and Planning Limitations}
\label{app:failure-modes}

We analyze when search and reobservation improve execution, then examine
candidate-ranking failures in a TwoRoom stress test.

\subsection{Separating Reobservation from Search}
\label{app:paired-recovery}
A paired comparison on PushT and Cube separates improvements from search
and recovery after reobservation. Both planners use the training-seed-0
checkpoint for each task. The resulting 300 pairs per distance
share weights, starts, goals, and execution budgets. ARCEM uses $T=0.2$
and scores generated commands before environment clipping.

\begin{figure}[!htb]
\centering
\includegraphics[width=\linewidth]{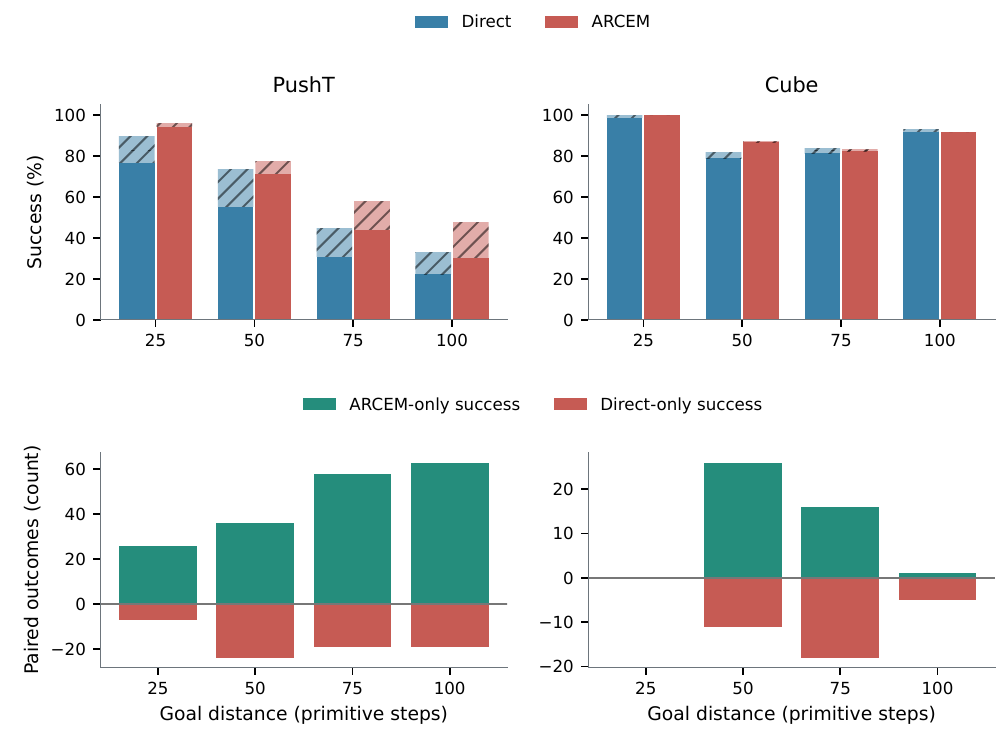}
\caption{\textbf{When reobservation and ARCEM help.}
Top: solid bars show first-stage success and hatched extensions show
additional success after reobservation. Within each distance, the left bar
is Direct and the right bar is ARCEM. Bottom: positive counts are
ARCEM-only successes and negative counts are Direct-only successes after
both stages. Each distance contains 300 paired evaluations.}
\label{fig:paired-recovery}
\end{figure}

For each planner, we count first-stage successes, additional second-stage
successes, and episodes that remain unsuccessful. Between planners, an
ARCEM-only success means that ARCEM succeeds within two stages while
Direct fails within both. A Direct-only success is the reverse event.
We count recovery after reobservation separately for each planner.
Figure~\ref{fig:paired-recovery} reports both decompositions.

On PushT, ARCEM succeeds on 183 evaluations where Direct fails and loses
69 Direct successes, for a net gain of 114 out of 1,200 evaluations.
At $D=25$, ARCEM has fewer second-stage rescues than Direct (5 versus 39)
because it already succeeds in 283 rather than 230 first-stage evaluations.
At $D=100$, first-stage success increases from 67 to 91, and the number
subsequently rescued increases from 32 to 52. Search can therefore improve
both the initial plan and the final outcome after reobservation.

Cube leaves less room for recovery. At $D=25$, Direct succeeds in 296
first-stage evaluations and rescues the remaining four. ARCEM succeeds in
all 300 in the first stage, so both finish at 100\%. Across all distances,
ARCEM succeeds in 43 cases where Direct fails, but fails in 34 cases where
Direct succeeds. It gains
15 successes at $D=50$ and loses two at $D=75$, while $D=25$ has equal
final success. At $D=75$, 16 evaluations improve and 18 regress, illustrating how a small aggregate change can conceal many
changes in individual outcomes.

ARCEM can replace a successful Direct plan with an unsuccessful one because
selection uses predicted latent costs.

\subsection{A Controlled Search Failure in TwoRoom}
\label{app:tworoom-failure}
A high-temperature stress test on TwoRoom examines a mismatch between
imagined and executable actions. We use $T=0.8$ at $D\in\{75,100\}$,
giving 1,800 evaluations. The mean fraction
of selected raw action components exceeding the environment's bounds is
19.75\%. Scoring the original commands while the environment clips them
can make a poor physical plan appear favorable.

For the first-stage diagnostic, we replay the candidate pools from all
search iterations, including the retained Direct plans, and verify that replay
reproduces the selected trajectories. Successful candidates exist for
1,775 of the 1,800 evaluations, but the selected plans succeed in only 878.
Among 667 cases where search loses a first-stage Direct success, 605 involve
wall collisions. This localizes much of the failure to candidate ranking
rather than absence of a successful proposal.

We then evaluate bounded-action scoring under the full two-stage protocol,
keeping temperature and search budget fixed. Candidate actions are clipped
in raw coordinates and renormalized for model input before scoring.
We rerun CEM with this score in every iteration, updating elite selection and
the residual distribution, while retaining the main evaluation's initial
and reobservation history inputs and all other settings.
Across the matched evaluations,
two-stage success rises from 78.11\% to 94.44\% at $D=75$ and from
69.33\% to 90.89\% at $D=100$.

\begin{figure}[!htb]
\centering
\includegraphics[width=\linewidth]{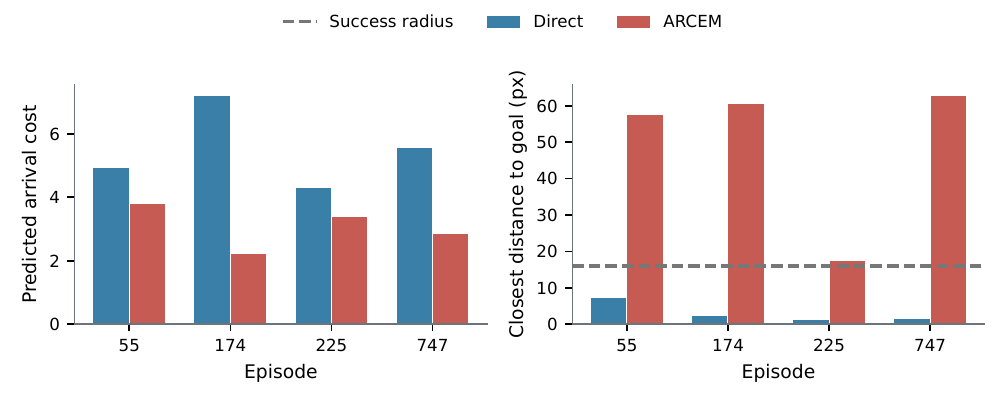}
\caption{\textbf{Search can discard a successful Direct plan.}
Predicted costs and closest replayed distances for four $D=75$ cases
where Direct succeeds and ARCEM at $T=0.8$ fails. Lower predicted cost
is preferred. Distances use complete first-plan replay without reobservation.
The dashed line marks the 16-pixel success radius.}
\label{fig:tworoom-failure}
\end{figure}

Figure~\ref{fig:tworoom-failure} compares predicted costs and replayed
distances for four cases where Direct succeeds but ARCEM fails.
For episode 55, search reduces predicted cost from $4.92$ to $3.80$,
yet its closest approach is $57.44$ pixels from the goal, outside the
$16$-pixel success radius. All four selected ARCEM plans have lower
predicted cost but worse physical proximity than Direct, with wall
collisions on 9--20 execution steps.

\subsection{Training and Deployment Conditions}
\label{app:planning-limits}
Training uses encoded observations at sampled chunk boundaries, whereas
planning feeds predicted latent states and generated chunk histories to
later actor calls. Student Forcing exposes the actor to generated
within-chunk action prefixes while keeping the latent state, intent,
preceding-chunk context, and expert targets fixed. Unlike DAgger
\citep{ross2011dagger}, it does not query an expert on learner-visited states.
Training on model-generated contexts and obtaining corrective targets
after physical state perturbations are complementary ways to address
these differences between training and deployment.

Chunk length controls imagined rollout resolution, while reobservation
provides physical feedback during execution. A future controller could
use model uncertainty to adapt either decision, for example by using
ensemble disagreement \citep{lakshminarayanan2016ensembles} to trigger
reobservation before the current plan ends. Comparisons with fixed
schedules could measure the trade-off between success, observation cost,
and planning cost.

\subsection{Qualitative Rollouts on PushT and Cube}
\label{app:qualitative-rollouts}
\begin{figure}[!htbp]
\centering
\includegraphics[width=\linewidth]{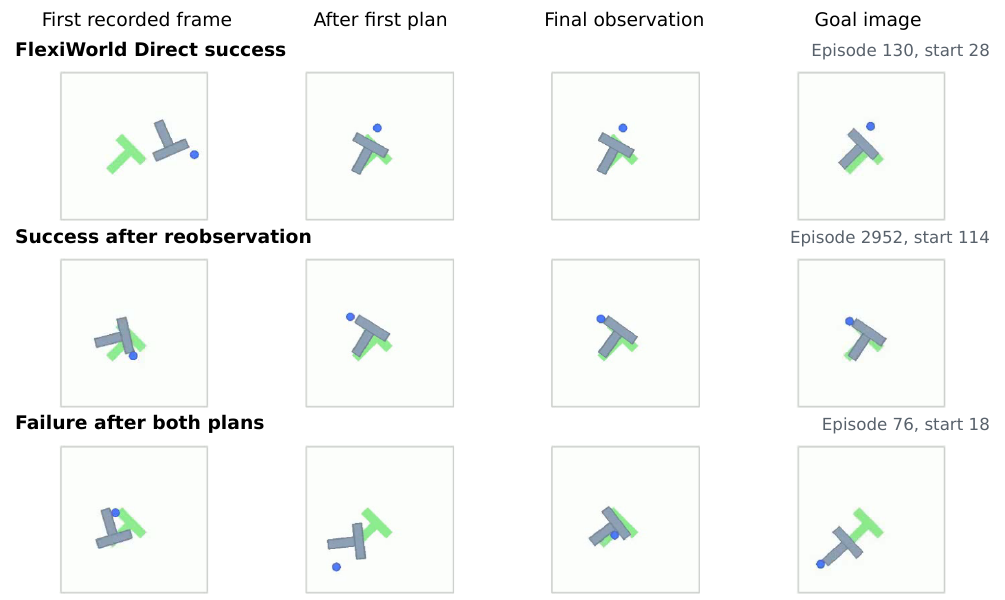}
\caption{\textbf{PushT: success, recovery, and remaining error.}
FlexiWorld Direct trajectories. The middle row places the object near its goal in the first plan, then
moves the agent closer to its target before success. The bottom row
remains unsuccessful after replanning. A trajectory that ends early
retains its terminal frame.}
\label{fig:pusht-cases}
\end{figure}

\begin{figure}[!htbp]
\centering
\includegraphics[width=\linewidth]{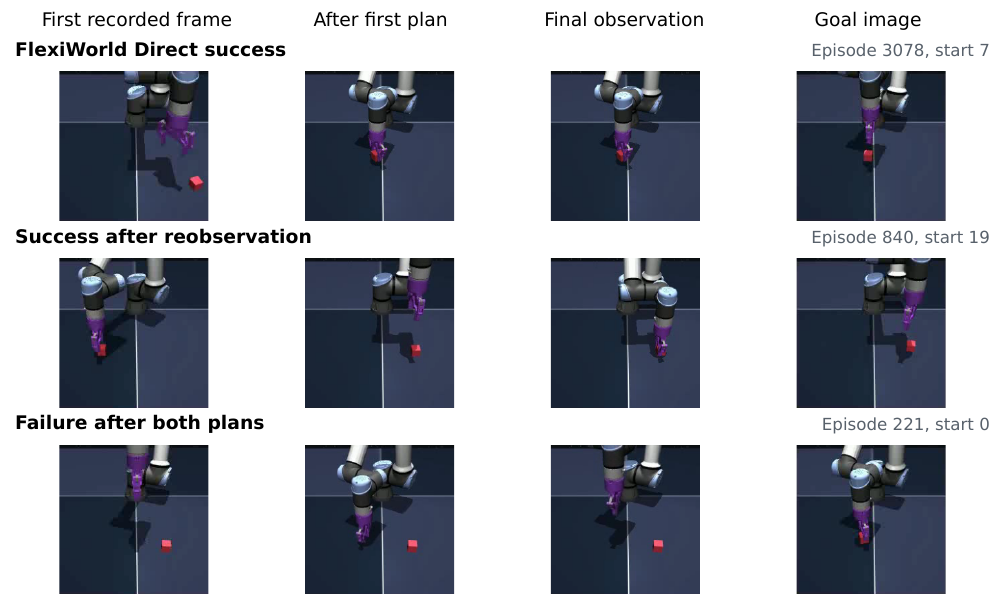}
\caption{\textbf{Cube: successful placement and incomplete recovery.}
FlexiWorld Direct trajectories. The middle row succeeds after reobservation,
while the bottom row leaves the cube away from its goal after both plans.
Terminal frames are held as in Figure~\ref{fig:pusht-cases}.}
\label{fig:cube-cases}
\end{figure}

Figures~\ref{fig:pusht-cases} and~\ref{fig:cube-cases} show success, recovery
after reobservation, and persistent failure at $D=75$ with $k=5$.
At replanning, the actor receives the last five executed action commands,
as in the quantitative evaluations.
For each task, we show the first recorded example in each category:
FlexiWorld Direct succeeds where INTACT fails, Direct succeeds after
reobservation, and Direct remains unsuccessful after both plans.
Frames after early termination repeat the final recorded state.

\clearpage

\clearpage


\clearpage
{\small
\begin{thebibliography}{44}
\providecommand{\natexlab}[1]{#1}
\providecommand{\url}[1]{\texttt{#1}}
\expandafter\ifx\csname urlstyle\endcsname\relax
  \providecommand{\doi}[1]{doi: #1}\else
  \providecommand{\doi}{doi: \begingroup \urlstyle{rm}\Url}\fi

\bibitem[Agarwal et~al.(2021)Agarwal, Schwarzer, Castro, Courville, and
  Bellemare]{agarwal2021evaluation}
Rishabh Agarwal, Max Schwarzer, Pablo~Samuel Castro, Aaron Courville, and
  Marc~G. Bellemare.
\newblock {Deep Reinforcement Learning at the Edge of the Statistical
  Precipice}, 2021.
\newblock URL \url{https://arxiv.org/abs/2108.13264}.

\bibitem[Andrychowicz et~al.(2017)Andrychowicz, Wolski, Ray, Schneider, Fong,
  Welinder, McGrew, Tobin, Abbeel, and Zaremba]{andrychowicz2017her}
Marcin Andrychowicz, Filip Wolski, Alex Ray, Jonas Schneider, Rachel Fong,
  Peter Welinder, Bob McGrew, Josh Tobin, Pieter Abbeel, and Wojciech Zaremba.
\newblock {Hindsight Experience Replay}, 2017.
\newblock URL \url{https://arxiv.org/abs/1707.01495}.

\bibitem[Assran et~al.(2023)Assran, Duval, Misra, Bojanowski, Vincent, Rabbat,
  LeCun, and Ballas]{assran2023ijepa}
Mahmoud Assran, Quentin Duval, Ishan Misra, Piotr Bojanowski, Pascal Vincent,
  Michael Rabbat, Yann LeCun, and Nicolas Ballas.
\newblock {Self-Supervised Learning from Images with a Joint-Embedding
  Predictive Architecture}, 2023.
\newblock URL \url{https://arxiv.org/abs/2301.08243}.

\bibitem[Assran et~al.(2025)Assran, Bardes, Fan, Garrido, Howes, Komeili,
  Muckley, Rizvi, Roberts, Sinha, Zholus, Arnaud, Gejji, Martin, Hogan, Dugas,
  Bojanowski, Khalidov, Labatut, Massa, Szafraniec, Krishnakumar, Li, Ma,
  Chandar, Meier, LeCun, Rabbat, and Ballas]{assran2025vjepa2}
Mido Assran, Adrien Bardes, David Fan, Quentin Garrido, Russell Howes, Mojtaba
  Komeili, Matthew Muckley, Ammar Rizvi, Claire Roberts, Koustuv Sinha, Artem
  Zholus, Sergio Arnaud, Abha Gejji, Ada Martin, Francois~Robert Hogan, Daniel
  Dugas, Piotr Bojanowski, Vasil Khalidov, Patrick Labatut, Francisco Massa,
  Marc Szafraniec, Kapil Krishnakumar, Yong Li, Xiaodong Ma, Sarath Chandar,
  Franziska Meier, Yann LeCun, Michael Rabbat, and Nicolas Ballas.
\newblock {V-JEPA 2: Self-Supervised Video Models Enable Understanding,
  Prediction and Planning}, 2025.
\newblock URL \url{https://arxiv.org/abs/2506.09985}.

\bibitem[Balestriero \& LeCun(2025)Balestriero and
  LeCun]{balestriero2025lejepa}
Randall Balestriero and Yann LeCun.
\newblock {LeJEPA: Provable and Scalable Self-Supervised Learning Without the
  Heuristics}, 2025.
\newblock URL \url{https://arxiv.org/abs/2511.08544}.

\bibitem[Bardes et~al.(2024)Bardes, Garrido, Ponce, Chen, Rabbat, LeCun,
  Assran, and Ballas]{bardes2024vjepa}
Adrien Bardes, Quentin Garrido, Jean Ponce, Xinlei Chen, Michael Rabbat, Yann
  LeCun, Mahmoud Assran, and Nicolas Ballas.
\newblock {Revisiting Feature Prediction for Learning Visual Representations
  from Video}, 2024.
\newblock URL \url{https://arxiv.org/abs/2404.08471}.

\bibitem[Bengio et~al.(2015)Bengio, Vinyals, Jaitly, and
  Shazeer]{bengio2015scheduledsampling}
Samy Bengio, Oriol Vinyals, Navdeep Jaitly, and Noam Shazeer.
\newblock Scheduled sampling for sequence prediction with recurrent neural
  networks.
\newblock \emph{arXiv preprint arXiv:1506.03099}, 2015.
\newblock \doi{10.48550/arXiv.1506.03099}.
\newblock URL \url{https://arxiv.org/abs/1506.03099}.

\bibitem[Chua et~al.(2018)Chua, Calandra, McAllister, and Levine]{chua2018pets}
Kurtland Chua, Roberto Calandra, Rowan McAllister, and Sergey Levine.
\newblock {Deep Reinforcement Learning in a Handful of Trials using
  Probabilistic Dynamics Models}, 2018.
\newblock URL \url{https://arxiv.org/abs/1805.12114}.

\bibitem[Dosovitskiy et~al.(2021)Dosovitskiy, Beyer, Kolesnikov, Weissenborn,
  Zhai, Unterthiner, Dehghani, Minderer, Heigold, Gelly, Uszkoreit, and
  Houlsby]{dosovitskiy2020vit}
Alexey Dosovitskiy, Lucas Beyer, Alexander Kolesnikov, Dirk Weissenborn,
  Xiaohua Zhai, Thomas Unterthiner, Mostafa Dehghani, Matthias Minderer, Georg
  Heigold, Sylvain Gelly, Jakob Uszkoreit, and Neil Houlsby.
\newblock {An Image is Worth 16x16 Words: Transformers for Image Recognition at
  Scale}.
\newblock In \emph{International Conference on Learning Representations}, 2021.
\newblock URL \url{https://openreview.net/forum?id=YicbFdNTTy}.

\bibitem[Du et~al.(2026)Du, Zhang, Wang, and Wang]{du2026vlwm}
Tianqi Du, Qi~Zhang, Yifei Wang, and Yisen Wang.
\newblock Beyond the next step: Variable-length latent world models for
  long-horizon planning.
\newblock \emph{arXiv preprint arXiv:2606.21775}, 2026.
\newblock \doi{10.48550/arXiv.2606.21775}.
\newblock URL \url{https://arxiv.org/abs/2606.21775}.

\bibitem[Gao \& Xu(2026)Gao and Xu]{gao2026fastleworldmodel}
Yuntian Gao and Xiangyu Xu.
\newblock Fast {LeWorldModel}.
\newblock \emph{arXiv preprint arXiv:2606.26217}, 2026.
\newblock \doi{10.48550/arXiv.2606.26217}.
\newblock URL \url{https://arxiv.org/abs/2606.26217}.

\bibitem[Ghosh et~al.(2021)Ghosh, Gupta, Reddy, Fu, Devin, Eysenbach, and
  Levine]{ghosh2019gcsl}
Dibya Ghosh, Abhishek Gupta, Ashwin Reddy, Justin Fu, Coline Devin, Benjamin
  Eysenbach, and Sergey Levine.
\newblock {Learning to Reach Goals via Iterated Supervised Learning}, 2021.
\newblock URL \url{https://arxiv.org/abs/1912.06088}.

\bibitem[Ha \& Schmidhuber(2018)Ha and Schmidhuber]{ha2018worldmodels}
David Ha and J{\"u}rgen Schmidhuber.
\newblock {World Models}, 2018.
\newblock URL \url{https://arxiv.org/abs/1803.10122}.

\bibitem[Hafner et~al.(2019)Hafner, Lillicrap, Fischer, Villegas, Ha, Lee, and
  Davidson]{hafner2019planet}
Danijar Hafner, Timothy Lillicrap, Ian Fischer, Ruben Villegas, David Ha,
  Honglak Lee, and James Davidson.
\newblock {Learning Latent Dynamics for Planning from Pixels}, 2019.
\newblock URL \url{https://arxiv.org/abs/1811.04551}.

\bibitem[Hafner et~al.(2020)Hafner, Lillicrap, Ba, and
  Norouzi]{hafner2019dreamer}
Danijar Hafner, Timothy Lillicrap, Jimmy Ba, and Mohammad Norouzi.
\newblock {Dream to Control: Learning Behaviors by Latent Imagination}, 2020.
\newblock URL \url{https://arxiv.org/abs/1912.01603}.

\bibitem[Hafner et~al.(2025)Hafner, Pasukonis, Ba, and
  Lillicrap]{hafner2023dreamerv3}
Danijar Hafner, Jurgis Pasukonis, Jimmy Ba, and Timothy Lillicrap.
\newblock {Mastering diverse control tasks through world models}.
\newblock \emph{Nature}, 640:\penalty0 647--653, 2025.
\newblock \doi{10.1038/s41586-025-08744-2}.
\newblock URL \url{https://doi.org/10.1038/s41586-025-08744-2}.

\bibitem[Hansen et~al.(2022)Hansen, Wang, and Su]{hansen2022tdmpc}
Nicklas Hansen, Xiaolong Wang, and Hao Su.
\newblock {Temporal Difference Learning for Model Predictive Control}, 2022.
\newblock URL \url{https://arxiv.org/abs/2203.04955}.

\bibitem[Hansen et~al.(2024)Hansen, Su, and Wang]{hansen2023tdmpc2}
Nicklas Hansen, Hao Su, and Xiaolong Wang.
\newblock {TD-MPC2: Scalable, Robust World Models for Continuous Control}.
\newblock In \emph{International Conference on Learning Representations}, 2024.
\newblock URL \url{https://openreview.net/forum?id=Oxh5CstDJU}.

\bibitem[Henderson et~al.(2018)Henderson, Islam, Bachman, Pineau, Precup, and
  Meger]{henderson2017deeprl}
Peter Henderson, Riashat Islam, Philip Bachman, Joelle Pineau, Doina Precup,
  and David Meger.
\newblock {Deep Reinforcement Learning that Matters}, 2018.
\newblock URL \url{https://arxiv.org/abs/1709.06560}.

\bibitem[Johannink et~al.(2018)Johannink, Bahl, Nair, Luo, Kumar, Loskyll,
  Ojea, Solowjow, and Levine]{johannink2018residual}
Tobias Johannink, Shikhar Bahl, Ashvin Nair, Jianlan Luo, Avinash Kumar,
  Matthias Loskyll, Juan~Aparicio Ojea, Eugen Solowjow, and Sergey Levine.
\newblock {Residual Reinforcement Learning for Robot Control}, 2018.
\newblock URL \url{https://arxiv.org/abs/1812.03201}.

\bibitem[Kornblith et~al.(2019)Kornblith, Norouzi, Lee, and
  Hinton]{kornblith2019cka}
Simon Kornblith, Mohammad Norouzi, Honglak Lee, and Geoffrey Hinton.
\newblock {Similarity of Neural Network Representations Revisited}, 2019.
\newblock URL \url{https://arxiv.org/abs/1905.00414}.

\bibitem[Lakshminarayanan et~al.(2017)Lakshminarayanan, Pritzel, and
  Blundell]{lakshminarayanan2016ensembles}
Balaji Lakshminarayanan, Alexander Pritzel, and Charles Blundell.
\newblock {Simple and Scalable Predictive Uncertainty Estimation using Deep
  Ensembles}, 2017.
\newblock URL \url{https://arxiv.org/abs/1612.01474}.

\bibitem[Liang et~al.(2026)Liang, Wang, Wang, Wang, Peng, Chen, Chua, and
  Vadakkepat]{liang2026adaptivechunking}
Yuanchang Liang, Xiaobo Wang, Kai Wang, Shuo Wang, Xiaojiang Peng, Haoyu Chen,
  David Kim~Huat Chua, and Prahlad Vadakkepat.
\newblock Adaptive action chunking at inference-time for vision-language-action
  models.
\newblock In \emph{Proceedings of the IEEE/CVF Conference on Computer Vision
  and Pattern Recognition}, 2026.
\newblock URL \url{https://arxiv.org/abs/2604.04161}.

\bibitem[Liu et~al.(2025)Liu, Hamid, Xie, Lee, Du, and
  Finn]{liu2025bidirectional}
Yuejiang Liu, Jubayer~Ibn Hamid, Annie Xie, Yoonho Lee, Max Du, and Chelsea
  Finn.
\newblock Bidirectional decoding: Improving action chunking via guided
  test-time sampling.
\newblock In \emph{International Conference on Learning Representations}, 2025.
\newblock URL \url{https://arxiv.org/abs/2408.17355}.

\bibitem[Loshchilov \& Hutter(2019)Loshchilov and Hutter]{loshchilov2017adamw}
Ilya Loshchilov and Frank Hutter.
\newblock {Decoupled Weight Decay Regularization}, 2019.
\newblock URL \url{https://arxiv.org/abs/1711.05101}.

\bibitem[Maes et~al.(2026)Maes, Le~Lidec, Scieur, LeCun, and
  Balestriero]{maes2026leworldmodel}
Lucas Maes, Quentin Le~Lidec, Damien Scieur, Yann LeCun, and Randall
  Balestriero.
\newblock {LeWorldModel}: Stable end-to-end joint-embedding predictive
  architecture from pixels.
\newblock \emph{arXiv preprint arXiv:2603.19312}, 2026.
\newblock \doi{10.48550/arXiv.2603.19312}.
\newblock URL \url{https://arxiv.org/abs/2603.19312}.

\bibitem[Nguyen et~al.(2026)Nguyen, Xu, and Huang]{nguyen2026latentgeometry}
Hoang Nguyen, Xiaohao Xu, and Xiaonan Huang.
\newblock Latent geometry beyond search: Amortizing planning in world models.
\newblock \emph{arXiv preprint arXiv:2605.08732}, 2026.
\newblock \doi{10.48550/arXiv.2605.08732}.
\newblock URL \url{https://arxiv.org/abs/2605.08732}.

\bibitem[Oquab et~al.(2024)Oquab, Darcet, Moutakanni, Vo, Szafraniec, Khalidov,
  Fernandez, Haziza, Massa, El-Nouby, Assran, Ballas, Galuba, Howes, Huang, Li,
  Misra, Rabbat, Sharma, Synnaeve, Xu, Jegou, Mairal, Labatut, Joulin, and
  Bojanowski]{oquab2023dinov2}
Maxime Oquab, Timothée Darcet, Théo Moutakanni, Huy Vo, Marc Szafraniec,
  Vasil Khalidov, Pierre Fernandez, Daniel Haziza, Francisco Massa, Alaaeldin
  El-Nouby, Mahmoud Assran, Nicolas Ballas, Wojciech Galuba, Russell Howes,
  Po-Yao Huang, Shang-Wen Li, Ishan Misra, Michael Rabbat, Vasu Sharma, Gabriel
  Synnaeve, Hu~Xu, Hervé Jegou, Julien Mairal, Patrick Labatut, Armand Joulin,
  and Piotr Bojanowski.
\newblock {DINOv2: Learning Robust Visual Features without Supervision}, 2024.
\newblock URL \url{https://arxiv.org/abs/2304.07193}.

\bibitem[Park et~al.(2025)Park, Frans, Eysenbach, and Levine]{park2025ogbench}
Seohong Park, Kevin Frans, Benjamin Eysenbach, and Sergey Levine.
\newblock {OGBench: Benchmarking Offline Goal-Conditioned RL}, 2025.
\newblock URL \url{https://arxiv.org/abs/2410.20092}.

\bibitem[Paszke et~al.(2019)Paszke, Gross, Massa, Lerer, Bradbury, Chanan,
  Killeen, Lin, Gimelshein, Antiga, Desmaison, Köpf, Yang, DeVito, Raison,
  Tejani, Chilamkurthy, Steiner, Fang, Bai, and Chintala]{paszke2019pytorch}
Adam Paszke, Sam Gross, Francisco Massa, Adam Lerer, James Bradbury, Gregory
  Chanan, Trevor Killeen, Zeming Lin, Natalia Gimelshein, Luca Antiga, Alban
  Desmaison, Andreas Köpf, Edward Yang, Zach DeVito, Martin Raison, Alykhan
  Tejani, Sasank Chilamkurthy, Benoit Steiner, Lu~Fang, Junjie Bai, and Soumith
  Chintala.
\newblock {PyTorch: An Imperative Style, High-Performance Deep Learning
  Library}, 2019.
\newblock URL \url{https://arxiv.org/abs/1912.01703}.

\bibitem[Pinneri et~al.(2020)Pinneri, Sawant, Blaes, Achterhold, Stueckler,
  Rolinek, and Martius]{pinneri2020icem}
Cristina Pinneri, Shambhuraj Sawant, Sebastian Blaes, Jan Achterhold, Joerg
  Stueckler, Michal Rolinek, and Georg Martius.
\newblock {Sample-efficient Cross-Entropy Method for Real-time Planning}, 2020.
\newblock URL \url{https://arxiv.org/abs/2008.06389}.

\bibitem[Rakhimov et~al.(2026)Rakhimov, Bredis, Maksyuta, and
  Gavrilov]{rakhimov2026qantara}
Ruslan Rakhimov, George Bredis, Yuriy Maksyuta, and Daniil Gavrilov.
\newblock {Qantara}: Bridge-flow training for multi-paradigm {JEPA} control.
\newblock \emph{arXiv preprint arXiv:2607.04978}, 2026.
\newblock \doi{10.48550/arXiv.2607.04978}.
\newblock URL \url{https://arxiv.org/abs/2607.04978}.

\bibitem[Ross et~al.(2011)Ross, Gordon, and Bagnell]{ross2011dagger}
Stephane Ross, Geoffrey~J. Gordon, and J.~Andrew Bagnell.
\newblock {A Reduction of Imitation Learning and Structured Prediction to
  No-Regret Online Learning}, 2011.
\newblock URL \url{https://arxiv.org/abs/1011.0686}.

\bibitem[Rubinstein(1999)]{rubinstein1999cem}
Reuven Rubinstein.
\newblock {The Cross-Entropy Method for Combinatorial and Continuous
  Optimization}.
\newblock \emph{Methodology and Computing in Applied Probability}, 1\penalty0
  (2):\penalty0 127--190, 1999.
\newblock \doi{10.1023/A:1010091220143}.
\newblock URL \url{https://doi.org/10.1023/A:1010091220143}.

\bibitem[Shin et~al.(2026)Shin, Chae, Kim, Park, and
  Sung]{shin2026adaptivechunking}
Yongjae Shin, Jongseong Chae, Seongmin Kim, Jongeui Park, and Youngchul Sung.
\newblock Adaptive action chunking via multi-chunk {Q} value estimation.
\newblock \emph{arXiv preprint arXiv:2605.10044}, 2026.
\newblock \doi{10.48550/arXiv.2605.10044}.
\newblock URL \url{https://arxiv.org/abs/2605.10044}.

\bibitem[Sun et~al.(2026)Sun, Zhao, and Zhang]{sun2026intact}
Junhan Sun, Hao Zhao, and Guofeng Zhang.
\newblock {INTACT}: Isomorphic intent-to-action learning for search-free world
  models.
\newblock \emph{arXiv preprint arXiv:2607.26056}, 2026.
\newblock \doi{10.48550/arXiv.2607.26056}.
\newblock URL \url{https://arxiv.org/abs/2607.26056}.

\bibitem[Vaswani et~al.(2017)Vaswani, Shazeer, Parmar, Uszkoreit, Jones, Gomez,
  Kaiser, and Polosukhin]{vaswani2017attention}
Ashish Vaswani, Noam Shazeer, Niki Parmar, Jakob Uszkoreit, Llion Jones,
  Aidan~N. Gomez, Lukasz Kaiser, and Illia Polosukhin.
\newblock {Attention Is All You Need}, 2017.
\newblock URL \url{https://arxiv.org/abs/1706.03762}.

\bibitem[Wang \& Ba(2019)Wang and Ba]{wang2019poplin}
Tingwu Wang and Jimmy Ba.
\newblock Exploring model-based planning with policy networks.
\newblock \emph{arXiv preprint arXiv:1906.08649}, 2019.
\newblock \doi{10.48550/arXiv.1906.08649}.
\newblock URL \url{https://arxiv.org/abs/1906.08649}.

\bibitem[Wang et~al.(2026)Wang, Xia, Zhou, Hu, Shi, Du, and Ye]{wang2026prism}
Yuhai Wang, Jiawei Xia, Rongxuan Zhou, Xiao Hu, Yongliang Shi, Jing Du, and
  Yang Ye.
\newblock {PRISM}: {PR}ior-guided imagination sampling in world models.
\newblock \emph{arXiv preprint arXiv:2606.07974}, 2026.
\newblock \doi{10.48550/arXiv.2606.07974}.
\newblock URL \url{https://arxiv.org/abs/2606.07974}.

\bibitem[Zhang et~al.(2026)Zhang, Terver, Zholus, Chitnis, Sutaria, Assran,
  Balestriero, Bar, Bardes, LeCun, and Ballas]{zhang2026hwm}
Wancong Zhang, Basile Terver, Artem Zholus, Soham Chitnis, Harsh Sutaria, Mido
  Assran, Randall Balestriero, Amir Bar, Adrien Bardes, Yann LeCun, and Nicolas
  Ballas.
\newblock Hierarchical planning with latent world models.
\newblock \emph{arXiv preprint arXiv:2604.03208}, 2026.
\newblock \doi{10.48550/arXiv.2604.03208}.
\newblock URL \url{https://arxiv.org/abs/2604.03208}.

\bibitem[Zhang et~al.(2025)Zhang, Liu, Chang, Schramm, and
  Boularias]{zhang2025arp}
Xinyu Zhang, Yuhan Liu, Haonan Chang, Liam Schramm, and Abdeslam Boularias.
\newblock Autoregressive action sequence learning for robotic manipulation.
\newblock \emph{IEEE Robotics and Automation Letters}, 10\penalty0
  (5):\penalty0 4898--4905, 2025.
\newblock \doi{10.1109/LRA.2025.3550849}.
\newblock URL \url{https://doi.org/10.1109/LRA.2025.3550849}.

\bibitem[Zhao et~al.(2026)Zhao, Nie, Lin, Luo, Gu, Fan, and
  Zeng]{zhao2026subjepa}
Kai Zhao, Dongliang Nie, Yuchen Lin, Zhehan Luo, Yixiao Gu, Deng-Ping Fan, and
  Dan Zeng.
\newblock {Sub-JEPA}: Subspace gaussian regularization for stable end-to-end
  world models.
\newblock \emph{arXiv preprint arXiv:2605.09241}, 2026.
\newblock \doi{10.48550/arXiv.2605.09241}.
\newblock URL \url{https://arxiv.org/abs/2605.09241}.

\bibitem[Zhao et~al.(2023)Zhao, Kumar, Levine, and Finn]{zhao2023act}
Tony~Z. Zhao, Vikash Kumar, Sergey Levine, and Chelsea Finn.
\newblock Learning fine-grained bimanual manipulation with low-cost hardware.
\newblock \emph{arXiv preprint arXiv:2304.13705}, 2023.
\newblock \doi{10.48550/arXiv.2304.13705}.
\newblock URL \url{https://arxiv.org/abs/2304.13705}.

\bibitem[Zhou et~al.(2024)Zhou, Pan, LeCun, and Pinto]{zhou2024dinowm}
Gaoyue Zhou, Hengkai Pan, Yann LeCun, and Lerrel Pinto.
\newblock {DINO-WM}: World models on pre-trained visual features enable
  zero-shot planning.
\newblock \emph{arXiv preprint arXiv:2411.04983}, 2024.
\newblock \doi{10.48550/arXiv.2411.04983}.
\newblock URL \url{https://arxiv.org/abs/2411.04983}.

\end{thebibliography}


}
\end{document}